\documentclass[10pt,twocolumn,letterpaper]{article}
\usepackage{cvpr}
\usepackage{times}
\usepackage{epsfig}
\usepackage{graphicx}
\usepackage{amsmath}
\usepackage{amssymb}
\usepackage[algoruled,boxed,lined]{algorithm2e}
\usepackage{color}
\usepackage{subfig}
\usepackage{xifthen}
\usepackage{rotating}
\usepackage{mathtools}
\usepackage{nicefrac}
\usepackage{array}
\usepackage{mathtools}
\usepackage{flushend}
\usepackage[pagebackref=true,breaklinks=true,colorlinks,bookmarks=false]{hyperref}

\cvprfinalcopy

\def\cvprPaperID{7058}

\ifcvprfinal\fi

\newcommand{\todo}[1][]{
\ifthenelse{\isempty{#1}}
	 {\textcolor{red}{(TODO)} \marginpar{\textcolor{red}{$\star$}}}
   {\textcolor{red}{(TODO: \marginpar{\textcolor{red}{$\star$}} #1)}}
}

\begin{document}

\title{Dataless Model Selection with the Deep Frame Potential}

\author{Calvin Murdock\textsuperscript{1} \hspace{0.5cm} Simon Lucey\textsuperscript{1,2}\\
\textsuperscript{1}Carnegie Mellon University \hspace{0.5cm} \textsuperscript{2}Argo AI\\
{\tt\small \{cmurdock,slucey\}@cs.cmu.edu}
}

\maketitle
\thispagestyle{empty}


\begin{abstract}
Choosing a deep neural network architecture is a fundamental problem in applications that require balancing performance and parameter efficiency. Standard approaches rely on ad-hoc engineering or computationally expensive validation on a specific dataset. We instead attempt to quantify networks by their intrinsic capacity for unique and robust representations, enabling efficient architecture comparisons without requiring any data. Building upon theoretical connections between deep learning and sparse approximation, we propose the deep frame potential: a measure of coherence that is approximately related to representation stability but has minimizers that depend only on network structure. This provides a framework for jointly quantifying the contributions of architectural hyper-parameters such as depth, width, and skip connections. We validate its use as a criterion for model selection and demonstrate correlation with generalization error on a variety of common residual and densely connected network architectures.

\end{abstract}

\section{Introduction}

Deep neural networks have dominated nearly every benchmark within the field of computer
vision. While this modern influx of deep learning originally began with the task of 
large-scale image recognition~\cite{krizhevsky2012imagenet}, new datasets, loss functions, and 
network configurations have quickly expanded its scope to include a much wider range of applications.
Despite this, the underlying architectures used to learn effective image representations are
generally consistent across all of them. 
This can be seen 
through the community's quick adoption of the newest state-of-the-art deep networks 
from AlexNet~\cite{krizhevsky2012imagenet} to VGGNet~\cite{simonyan2014very}, 
ResNets~\cite{he2016deep}, DenseNets~\cite{huang2017densely}, and so on. But this begs the question: 
why do some deep network architectures work better than others? Despite years of groundbreaking
empirical results, an answer to this question still remains elusive. 

\begin{figure}
\centering
\captionsetup[subfigure]{labelformat=empty, justification=centering}
\subfloat[(a) Chain Network]{
\includegraphics[width=0.3\columnwidth]{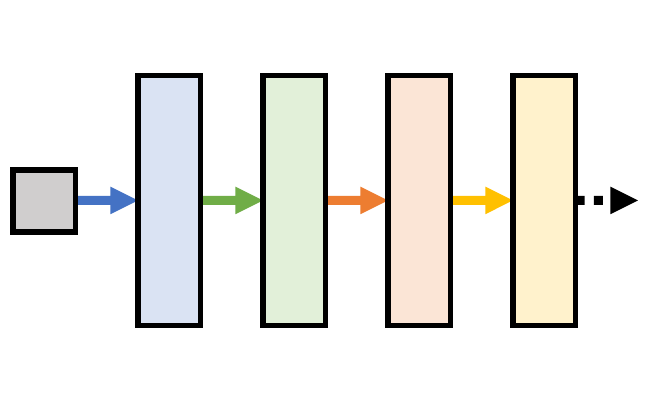}
}\hspace*{\fill}
\subfloat[(b) Residual Network]{
\includegraphics[width=0.3\columnwidth]{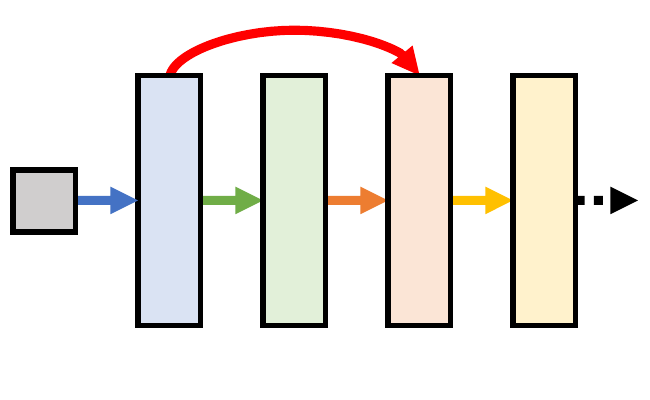}
}\hspace*{\fill}
\subfloat[(c) Densely Connected Convolutional Network]{
\includegraphics[width=0.3\columnwidth]{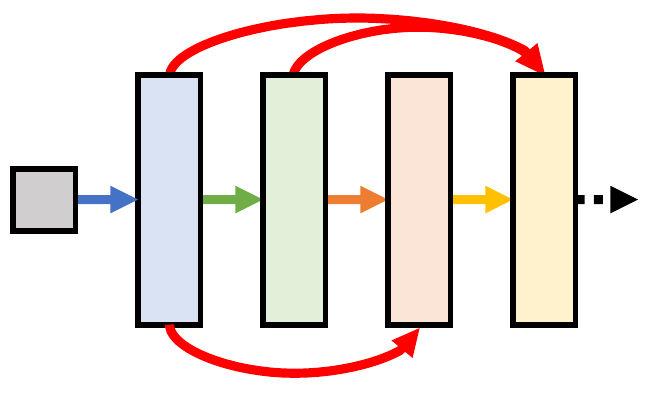}
}

\subfloat[(d) Induced Dictionary Structures for Sparse Approximation]{
\includegraphics[width=0.3\columnwidth]{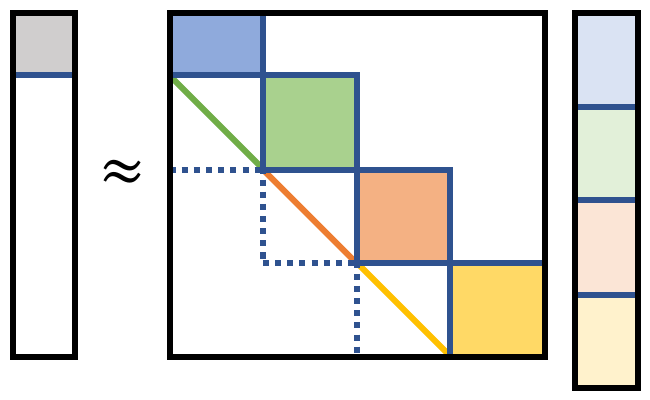}
\hspace*{0.4em}
\includegraphics[width=0.3\columnwidth]{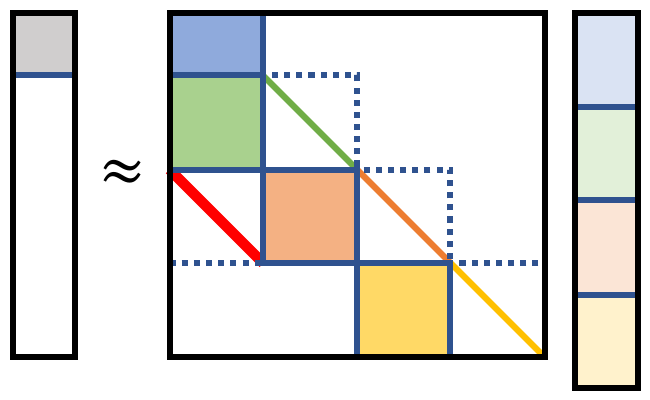}
\hspace*{0.4em}
\includegraphics[width=0.3\columnwidth]{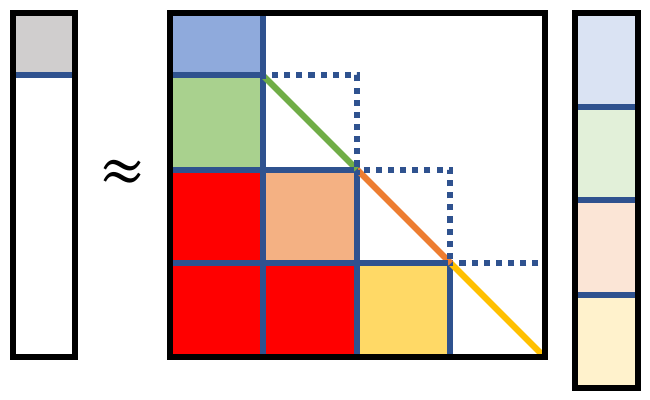}
}

\caption{
\small Why are some deep neural network architectures better than others? 
In comparison to (a) standard chain connections,
skip connections like those in (b) ResNets~\cite{he2016deep} and 
(c) DenseNets~\cite{huang2017densely} have demonstrated significant 
improvements in training effectiveness, parameter efficiency, and 
generalization performance. We provide one possible explanation for this
phenomenon by approximating network activations as (d) solutions
to sparse approximation problems with different induced
dictionary structures. 
}
\label{fig:front}
\end{figure}

Fundamentally, the difficulty in comparing network architectures arises from the lack of a 
theoretical foundation for characterizing their generalization capacities. 
Shallow machine learning techniques like support vector machines~\cite{cortes1995support} were 
aided by theoretical tools like the VC-dimension~\cite{vapnik1971uniform} for determining when 
their predictions could be trusted to avoid overfitting. Deep neural networks, 
on the other hand, have eschewed similar analyses due to their complexity. 
Theoretical explorations of deep network 
generalization~\cite{neyshabur2017exploring} are often disconnected from 
practical applications 
and rarely provide actionable insight into how architectural hyper-parameters contribute
to performance. 

Building upon recent connections between deep learning and sparse 
approximation~\cite{papyan2017convolutional, murdock2018deep}, we instead 
interpret feed-forward deep networks as algorithms for approximate inference in related 
sparse coding problems. These problems aim to optimally reconstruct zero-padded input images as
sparse, nonnegative linear combinations of atoms from architecture-dependent dictionaries, 
as shown in Fig.~\ref{fig:front}. 
We propose to indirectly analyze practical 
deep network architectures with complicated skip connections, like residual networks 
(ResNets)~\cite{he2016deep} and 
densely connected convolutional networks (DenseNets)~\cite{huang2017densely}, simply through the 
dictionary structures that they induce. 

To accomplish this, we introduce the deep frame potential for summarizing the 
interactions between parameters in feed-forward deep networks.
As a lower bound on mutual coherence--the maximum magnitude of the normalized inner products 
between all pairs of dictionary atoms~\cite{donoho2003optimally}--it is 
theoretically tied to generalization properties of the 
related sparse coding problems.
However, its minimizers depend only on the dictionary structures induced by   
the corresponding network architectures. This enables dataless model comparison by jointly quantifying 
contributions of depth, width, and connectivity.

Our approach is motivated by sparse approximation theory~\cite{elad2010sparse}, a field that encompasses 
properties like uniqueness and robustness of 
shallow, overcomplete representations. In sparse coding, capacity is controlled 
by the number of dictionary atoms used in sparse data reconstructions. While
more parameters  
allow for more accurate representations,
they may also increase input sensitivity for worse generalization performance.
Conceptually, this is comparable to overfitting in nearest-neighbor classification,
where representations are sparse, one-hot indicator vectors corresponding to nearest training examples.
As the number of training data increases, the distance between them decreases, so they are more likely to be 
confused with
one another. Similarly, nearby dictionary atoms may introduce instability that causes representations 
of similar data points to become very far apart leading to poor generalization performance.
Thus, there is a fundamental tradeoff between the capacity and robustness of shallow representations
due to the proximity of dictionary atoms as measured by mutual coherence.

\begin{figure}
\centering

\hspace*{\fill}
\subfloat[Chain Network Gram Matrix]{
\includegraphics[height=1.4cm]{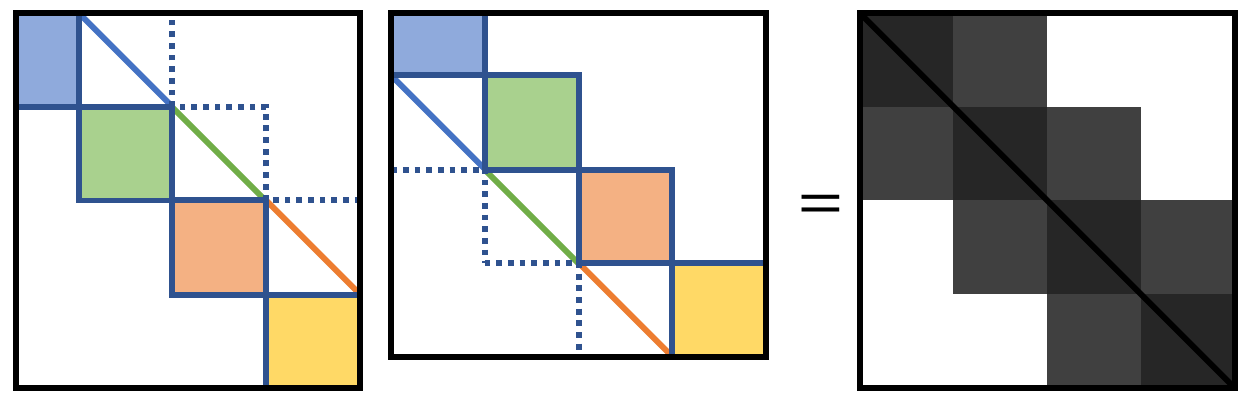}
}
\hspace*{\fill}
\subfloat[ResNet]{
\includegraphics[height=1.4cm]{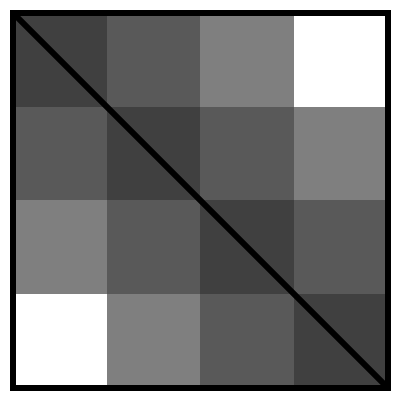}
}
\hspace*{\fill}
\subfloat[DenseNet]{
\includegraphics[height=1.4cm]{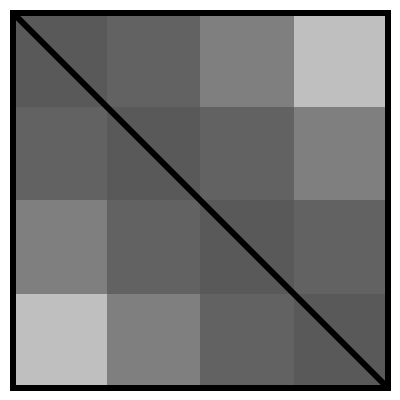}
}
\hspace*{\fill}

\subfloat[Minimum Deep Frame Potential]{
\includegraphics[width=0.48\columnwidth]{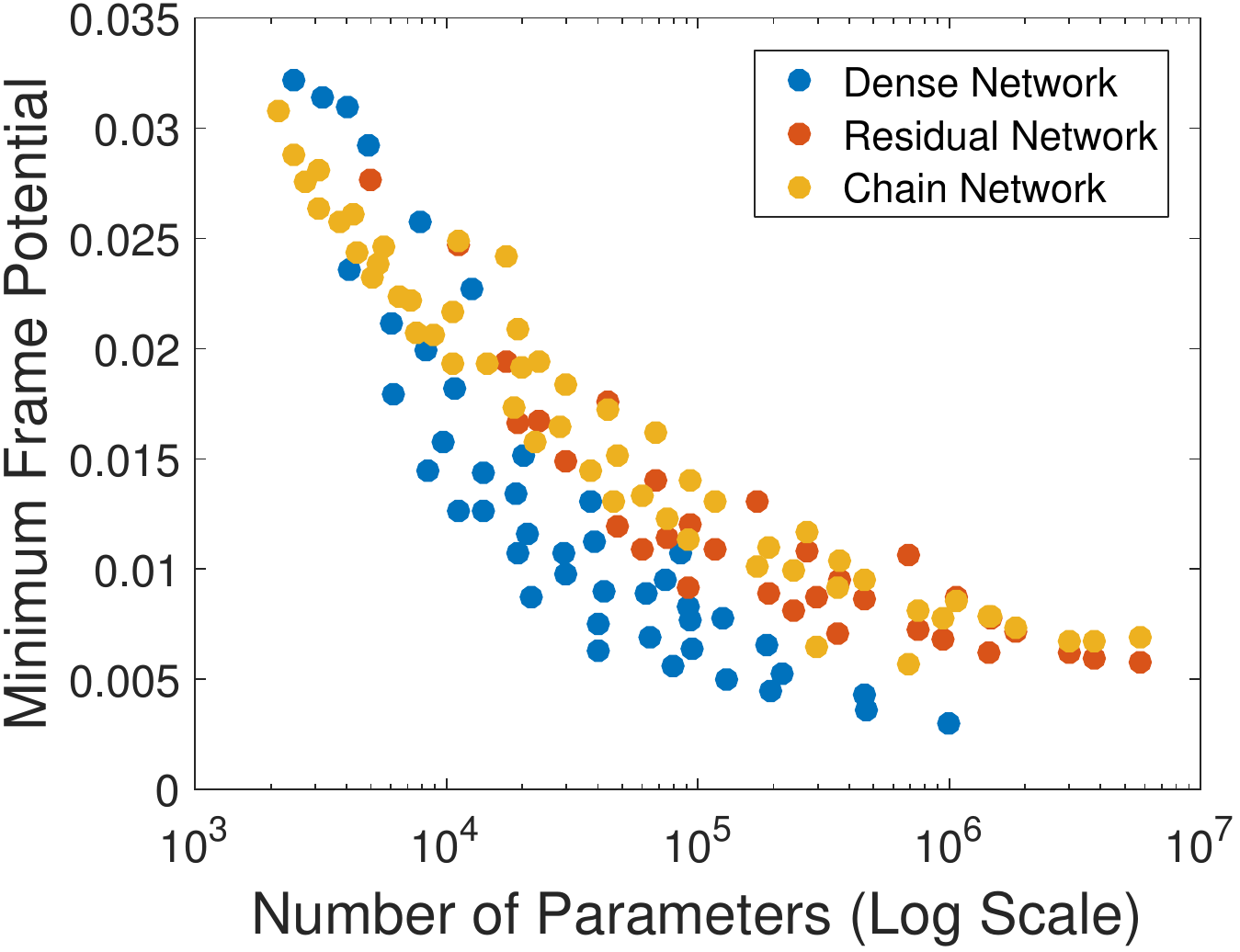}
}
\subfloat[Validation Error]{
\includegraphics[width=0.48\columnwidth]{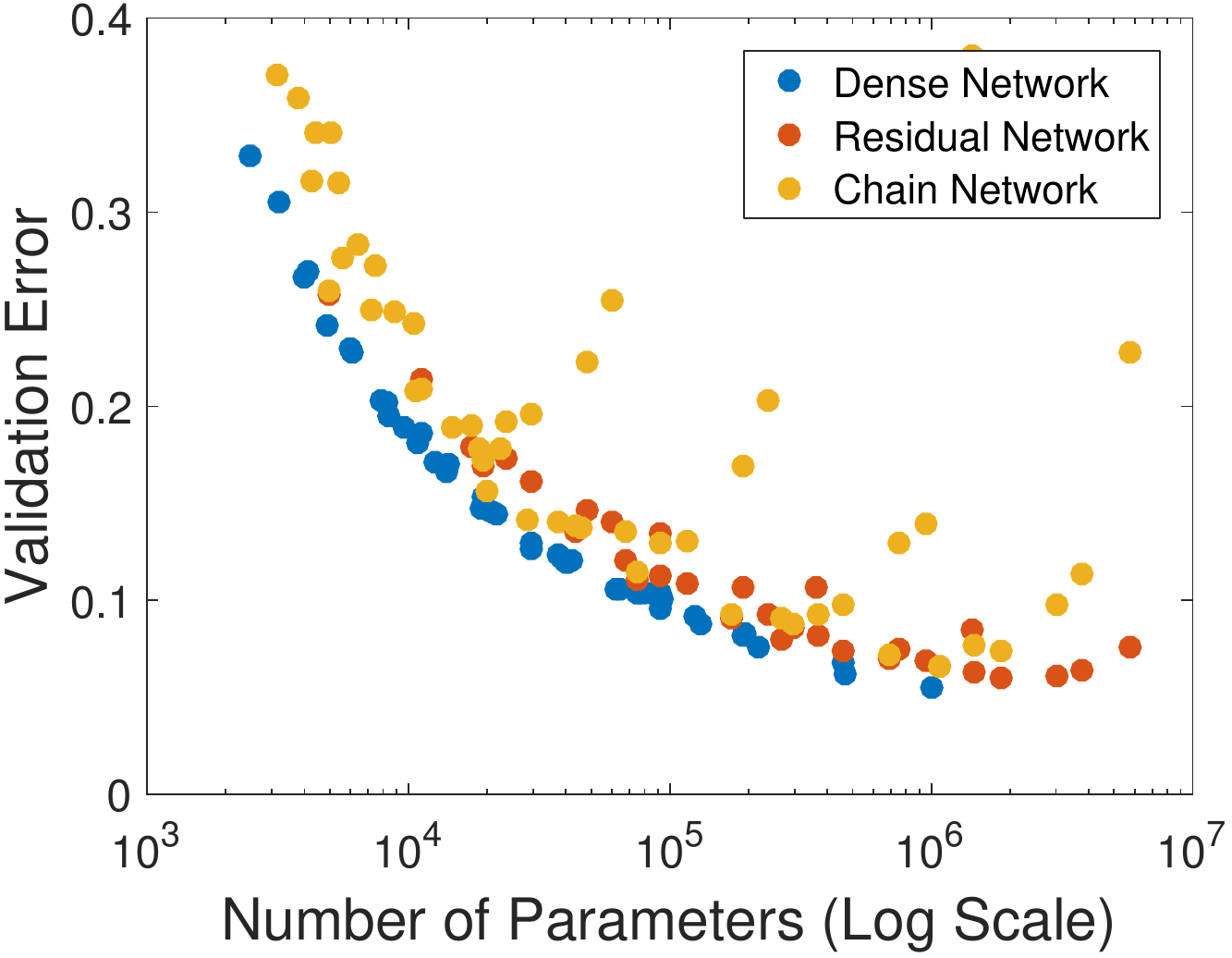}
}
\caption{
\small Parameter count is not a good indicator of generalization performance for deep networks. Instead, we 
compare different 
network architectures via the minimum deep frame potential, the 
average nonzero magnitude of inner products between atoms of architecture-induced 
dictionaries. In comparison to (a) chain networks, 
skip connections in (b) residual networks and (c) densely connected networks produce Gram matrix structures 
with more nonzero elements allowing for (d) lower deep frame potentials across network sizes. This correlates with improved parameter efficiency giving (e) lower validation error with fewer parameters.
}
\label{fig:gram}
\end{figure}

However, deep representations have not shown the same correlation between model size and 
sensitivity~\cite{zhang2017understanding}. 
While adding more layers to a deep neural network increases its capacity, it also simultaneously
introduces implicit regularization to reduce overfitting. 
This can be explained through the proposed connection to sparse coding, where additional layers
increase both capacity and effective input dimensionality.
In a higher-dimensional space, dictionary atoms can be spaced further apart for more robust representations.
Furthermore, architectures with denser skip 
connections induce dictionary structures with more nonzero elements, which provides additional 
freedom to further reduce mutual coherence with fewer parameters as shown in Fig.~\ref{fig:gram}.

We propose to use the minimum deep frame potential
as a cue for model selection. Instead of requiring expensive validation 
on a specific dataset to approximate generalization performance, architectures are  
chosen based on how efficiently they can reduce the minimum achievable mutual coherence with respect to the 
number of model parameters. In this paper, we provide an efficient frame potential minimization method for a 
general class of convolutional networks with skip connections, of which ResNets and DenseNets are shown to be 
special cases. Furthermore, we derive an analytic expression for the minimum value in the 
case of fully-connected chain networks. Experimentally, we 
demonstrate correlation with validation error across a variety of 
network architectures.

\section{Background and Related Work}

Due to the vast space of possible deep network architectures and the computational difficulty
in training them, deep model selection has largely been guided by ad-hoc engineering and 
human ingenuity. 
While progress slowed in the years following early breakthroughs~\cite{lecun1998gradient},
recent interest in deep learning architectures began anew due to empirical 
successes largely attributed to computational advances like efficient training using GPUs and 
rectified linear unit (ReLU) activation 
functions~\cite{krizhevsky2012imagenet}. Since then, numerous architectural changes have been
proposed. For example, much deeper networks with residual connections were shown 
to achieve consistently better performance with fewer parameters~\cite{he2016deep}.
Building upon this, densely connected convolutional networks with skip connections between 
more layers yielded
even better performance~\cite{huang2017densely}.
While theoretical explanations for these improvements were lacking, consistent experimentation 
on standardized benchmark datasets continued to drive empirical success.

However, due to slowing progress and the need for increased 
accessibility of deep learning techniques to a wider range of practitioners, more principled
approaches to architecture search have recently gained traction. 
Motivated by observations of extreme redundancy in the parameters of trained 
networks~\cite{denil2013predicting}, techniques have been proposed to systematically reduce
the number of parameters without adversely affecting performance. 
Examples include sparsity-inducing regularizers during training~\cite{alvarez2016learning}
or through post-processing to prune the parameters of trained networks~\cite{he2017channel}. 
Constructive approaches to model selection like neural architecture search~\cite{elsken2019neural}
instead attempt to compose 
architectures from basic building blocks through tools like reinforcement learning.  
Efficient model scaling has also been proposed to enable more effective 
grid search for selecting architectures subject to resource 
constraints~\cite{tan2019efficientnet}.
While automated techniques can match or even 
surpass manually engineered alternatives, they require a validation dataset 
and rarely provide insights transferable to other settings. 

To better understand the implicit benefits of different network architectures, there have 
been adjacent theoretical explorations of deep network generalization. These works are often 
motivated by the surprising observation that good performance can still be achieved using 
highly over-parametrized models 
with degrees of freedom that surpass the number of training 
data. This contradicts many commonly accepted ideas about 
generalization, spurning new experimental explorations
that have demonstrated properties unique to deep learning. Examples include the ability of 
deep networks to express random 
data labels~\cite{zhang2017understanding} with a tendency towards learning 
simple patterns first~\cite{arpit2017closer}. While exact theoretical explanations are lacking,
empirical measurements of network sensitivity such as the Jacobian norm have been shown to correlate
with generalization~\cite{novak2018sensitivity}. Similarly, Parseval 
regularization~\cite{moustapha2017parseval} encourages robustness by constraining the Lipschitz
constants of individual layers. 

Due to the difficulty in analyzing deep networks directly, other approaches have 
instead drawn connections to the rich field of sparse approximation theory.
The relationship between feed-forward neural networks and principal component analysis has long been 
known for the case of linear activations~\cite{baldi1989neural}.
More recently, nonlinear deep networks with ReLU activations 
have been linked to multilayer sparse coding 
to prove theoretical properties of deep 
representations~\cite{papyan2017convolutional}. This connection has been used to motivate
new recurrent architecture designs that resist adversarial noise attacks~\cite{romano2018adversarial},
improve classification performance~\cite{sulam2019multi},
or enforce prior knowledge through output constraints~\cite{murdock2018deep}. 

\section{Deep Learning as Sparse Approximation}

To derive our criterion for model selection, we build upon recent connections between deep neural 
networks and sparse approximation. Specifically, consider a feed-forward network 
$\boldsymbol{f}(\boldsymbol{x})=\phi_l(\mathbf{B}_l^\mathsf{T}\cdots\phi_1(\mathbf{B}_1^\mathsf{T}(\boldsymbol{x})))$ 
constructed as the composition of linear transformations with parameters $\mathbf{B}_j$
and nonlinear activation functions $\phi_j$. Equivalently, $\boldsymbol{f}(\boldsymbol{x})=\boldsymbol{w}_l$ 
where $\boldsymbol{w}_j=\mathbf{B}_j^\mathsf{T}\boldsymbol{w}_{j-1}$ are the layer activations 
for layers $j=1,\dotsc,l$ and $\boldsymbol{w}_0=\boldsymbol{x}$. In many modern state-of-the-art 
networks, the
ReLU activation function has been adopted due to its effectiveness and 
computational efficiency. It can also be interpreted as the nonnegative 
soft-thresholding 
proximal operator associated with the function $\Phi$ in Eq.~\ref{eq:prox}, a nonnegativity
constraint $\mathbb{I}(\boldsymbol{w}\geq\mathbf{0})$ and a sparsity-inducing $\ell_1$ 
penalty with a weight determined by the scalar bias parameter $\lambda$.
\begin{gather}
\Phi(\boldsymbol{w}) =\mathbb{I}(\boldsymbol{w}\geq\mathbf{0})+\lambda\left\lVert \boldsymbol{w}\right\rVert _{1}
\label{eq:prox}\\
\phi(\boldsymbol{x}) =\mathrm{ReLU}(\boldsymbol{x}-\lambda\mathbf{1})=\underset{\boldsymbol{w}}{\arg\min}\tfrac{1}{2}\left\lVert \boldsymbol{w}-\boldsymbol{x}\right\rVert _{2}^{2}+\Phi(\boldsymbol{w})\nonumber
\end{gather}

Thus, the forward pass of a deep network is equivalent to  
a layered thresholding pursuit algorithm for approximating the solution of a multi-layer sparse coding
model~\cite{papyan2017convolutional}. 
Results from shallow sparse approximation theory can then 
be adapted to bound the accuracy of this approximation, which improves as the mutual coherence 
defined below in Eq.~\ref{eq:mutual_coherence} decreases,
and indirectly analyze other theoretical properties 
of deep networks like uniqueness and robustness. 

\subsection{Sparse Approximation Theory}

Sparse approximation theory considers representations of data vectors 
$\boldsymbol{x}\in\mathbb{R}^d$ as sparse linear combinations
$\boldsymbol{x}\approx\sum_j w_j\boldsymbol{b}_j=\mathbf{B}\boldsymbol{w}$ 
of atoms from an over-complete dictionary $\mathbf{B}\in\mathbb{R}^{d\times k}$.
The number of atoms $k$ is greater than 
the dimensionality $d$ and the 
number of nonzero coefficients $\left\Vert\boldsymbol{w}\right\Vert_0$ in the representation $\boldsymbol{w}\in\mathbb{R}^k$ is small.

Through applications like compressed sensing~\cite{donoho2006compressed}, sparsity
has been found to exhibit theoretical properties that enable data representation with 
efficiency far greater than what was previously thought possible. 
Central to these results is the 
requirement that the dictionary be ``well-behaved,'' essentially ensuring
that its columns are not too similar. For undercomplete matrices with $k\leq d$, this is satisfied 
by enforcing orthogonality, but overcomplete dictionaries require other 
conditions. Specifically, we focus our attention on the mutual coherence $\mu$, 
the maximum magnitude normalized inner product of all pairs of dictionary atoms.
Equivalently, it is the maximum magnitude off-diagonal element in the Gram matrix 
$\mathbf{G}=\tilde{\mathbf{B}}^{\mathsf{T}}\tilde{\mathbf{B}}$ where the columns of $\tilde{\mathbf{B}}$
are normalized to have unit norm:
\begin{equation}
\mu=\max_{i\neq j}\frac{|\boldsymbol{b}_{i}^{\mathsf{T}}\boldsymbol{b}_{j}|}{\left\lVert \boldsymbol{b}_{i}\right\rVert \left\lVert \boldsymbol{b}_{j}\right\rVert }=\max_{i,j}|(\mathbf{G}-\mathbf{I})_{ij}|
\label{eq:mutual_coherence}
\end{equation}
\begin{minipage}{\columnwidth}
We are primarily motivated by the observation that a model's 
capacity for low mutual coherence 
increases along with its capacity for both  
memorizing more training data through unique 
representations and generalizing to more validation data through robustness to input perturbations.
\end{minipage} 

With an overcomplete dictionary, there is a space of coefficients $\boldsymbol{w}$ that can all exactly
reconstruct any data point as $\boldsymbol{x}=\mathbf{B}\boldsymbol{w}$, which would not support
discriminative representation learning. 
However, if representations from a mutually incoherent dictionary are sufficiently sparse, then 
they are guaranteed to be optimally sparse and unique~\cite{donoho2003optimally}. Specifically, if 
the number of nonzeros 
$\left\lVert \boldsymbol{w}\right\rVert _{0}<\tfrac{1}{2}(1+\mu^{-1})$, then $\boldsymbol{w}$ is the unique,
sparsest representation for $\boldsymbol{x}$. Furthermore, if 
$\left\lVert \boldsymbol{w}\right\rVert _{0}<(\sqrt{2}-0.5)\mu^{-1}$, then it can be found efficiently 
by convex optimization with $\ell_1$ regularization. Thus, minimizing the 
mutual coherence of a dictionary increases 
its capacity for uniquely representing data points.

Sparse representations are also robust to input 
perturbations~\cite{donoho2005stable}. Specifically, given a noisy datapoint
$\boldsymbol{x}=\boldsymbol{x}_{0}+\boldsymbol{z}$ where $\boldsymbol{x}_{0}$ 
can be represented exactly as $\boldsymbol{x}_{0}=\mathbf{B}\boldsymbol{w}_{0}$ with 
$\left\lVert \boldsymbol{w}_{0}\right\rVert _{0}\leq\tfrac{1}{4}\left(1+\mu^{-1}\right)$
and the noise $\boldsymbol{z}$ 
has bounded magnitude $\left\lVert \boldsymbol{z}\right\rVert _{2}\leq\epsilon$, then $\boldsymbol{w}_0$
can be approximated by solving the $\ell_{1}$-penalized LASSO problem:
\begin{equation}
\underset{\boldsymbol{w}}{\arg\min}\left\lVert \boldsymbol{x}-\mathbf{B}\boldsymbol{w}\right\rVert _{2}^{2}+\lambda\left\lVert \boldsymbol{w}\right\rVert _{1}
\label{eq:lasso}
\end{equation}
Its solution is stable and the approximation error is bounded from above in 
Eq.~\ref{eq:robustness}, where 
$\delta(\boldsymbol{x},\lambda)$ is a constant.
\begin{equation}
\left\lVert \boldsymbol{w}-\boldsymbol{w}_{0}\right\rVert _{2}^{2}\leq\frac{\left(\epsilon+\delta(\boldsymbol{x},\lambda)\right)^{2}}{1-\mu(4\left\Vert\boldsymbol{w}\right\Vert_0-1)}
\label{eq:robustness}
\end{equation}
Thus, minimizing the mutual coherence of a dictionary decreases the sensitivity of its sparse representations for improved robustness.
This is similar to evaluating input sensitivity using the Jacobian norm~\cite{novak2018sensitivity}. However, 
instead of estimating the average perturbation error over validation data, it bounds the 
worst-cast error over all possible data.

\subsection{Deep Component Analysis}

While deep representations can be analyzed by accumulating the effects of approximating 
individual layers in a chain network as shallow sparse coding
problems~\cite{papyan2017convolutional}, this strategy cannot be easily adapted to account for more 
complicated interactions between layers. 
Instead, we adapt the framework of Deep Component Analysis~\cite{murdock2018deep}, which jointly 
represents all layers in a neural network as the single sparse coding problem in
Eq.~\ref{eq:chain}. The ReLU activations $\boldsymbol{w}_j\in\mathbb{R}^{k_j}$ of a 
feed-forward chain network
approximate the solutions to a joint optimization problem where 
$\boldsymbol{w}_0=\boldsymbol{x}\in\mathbb{R}^{k_0}$ and the regularization functions $\Phi_j$ are 
nonnegative sparsity-inducing penalties as defined in Eq.~\ref{eq:prox}.
\begin{align}
\boldsymbol{w}_{j} & \coloneqq\phi_{j}(\mathbf{B}_{j}^{\mathsf{T}}\boldsymbol{w}_{j-1})\quad \forall j=1,\dotsc,l
\label{eq:chain}\\
 & \approx\underset{\{\boldsymbol{w}_{j}\}}{\arg\min}\sum_{j=1}^{l}\left\lVert \mathbf{B}_{j}\boldsymbol{w}_{j}-\boldsymbol{w}_{j-1}\right\rVert _{2}^{2}+\Phi_{j}(\boldsymbol{w}_{j})\nonumber
\end{align}
The compositional constraints between adjacent layers are relaxed and replaced 
by reconstruction error penalty terms, resulting in a convex, nonnegative sparse coding problem. 

By combining the terms in the summation of Eq.~\ref{eq:chain} together into a single 
system, this problem can be equivalently represented as shown in Eq.~\ref{eq:chain_opt}. The latent  
variables $\boldsymbol{w}_j$ for each layer are stacked  in the vector 
$\boldsymbol{w}$, the regularizer $\Phi(\boldsymbol{w})=\sum_j \Phi_j(\boldsymbol{w}_j)$, 
and the input $\boldsymbol{x}$
is augmented with zeros.
\begin{equation}
\hspace{-0.4em}
\arraycolsep=1.3pt\def\arraystretch{1.0}
\underset{\boldsymbol{w}}{\arg\min}\left\Vert \begin{array}{c}
\\
\\
\\
\\
\\
\end{array}\hspace{-0.5em}\right.\overbrace{\left[\begin{array}{cccc}
\mathbf{B}_{1} & \mathbf{0} &  & \\
-\mathbf{I} & \mathbf{B}_{2} & \ddots\\
 & \ddots & \ddots & \mathbf{0}\\
 &  & -\mathbf{I} & \mathbf{B}_{l}
\end{array}\right]}^{\mathbf{B}}\overbrace{\left[\begin{array}{c}
\boldsymbol{w}_{1}\\
\vphantom{\ddots}\boldsymbol{w}_{2}\\
\vdots\\
\boldsymbol{w}_{l}
\end{array}\right]}^{\boldsymbol{w}}-\left[\begin{array}{c}
\boldsymbol{x}\\
\vphantom{\ddots}\mathbf{0}\\
\vdots\\
\mathbf{0}
\end{array}\right]\left.\begin{array}{c}
\\
\\
\\
\\
\\
\end{array}\hspace{-0.3em}\right\Vert _{2}^{2}\hspace{-0.4em}+\Phi(\boldsymbol{w})
\hspace{-0.3em}
\label{eq:chain_opt}
\end{equation}
The layer parameters 
$\mathbf{B}_j\in\mathbb{R}^{k_{j-1}\times k_j}$ are blocks in the induced dictionary $\mathbf{B}$, 
which has $\sum_j k_{j-1}$ rows and $\sum_j k_j$ columns. It has a  
structure of nonzero elements that summarizes the corresponding feed-forward deep network architecture 
wherein the off-diagonal identity matrices are connections between adjacent layers. 

Model capacity can be increased both by adding 
parameters to a layer or by adding 
layers, which implicitly pads the input data $\boldsymbol{x}$ with more zeros. This can actually reduce 
the dictionary's mutual coherence because it increases the system's dimensionality. 
Thus, depth allows model complexity to scale jointly alongside effective input 
dimensionality so that the induced dictionary structures still have the capacity 
for low mutual coherence and improved capabilities for memorization and generalization.

\subsection{Architecture-Induced Dictionary Structure}

In this section, we extend this model formulation to incorporate  
more complicated network architectures. 
Because mutual coherence is dependent on normalized dictionary atoms, it
can be reduced by increasing the number of nonzero elements, which reduces
the magnitudes of the dictionary elements and their inner products. 
In Eq.~\ref{eq:dense_structure}, we replace the identity connections of Eq.~\ref{eq:chain_opt} 
with blocks of nonzero parameters to allow for lower mutual coherence.
\begin{equation}
\arraycolsep=1pt\def\arraystretch{1.0}
\mathbf{B}=\left[\begin{array}{cccc}
\mathbf{B}_{11} & \vphantom{\ddots}\mathbf{0} &  & \\
\mathbf{B}_{21}^{\mathsf{T}} & \mathbf{B}_{22} & \ddots &  \\
\vdots & \ddots & \ddots & \mathbf{0}\\
\mathbf{B}_{l1}^{\mathsf{T}} & \cdots & \mathbf{B}_{l(l-1)}^{\mathsf{T}} & \mathbf{B}_{ll}
\end{array}\right]
\label{eq:dense_structure}
\end{equation}
This structure is induced by the feed-forward activations in 
Eq.~\ref{eq:dense}, which again approximate the solutions to a nonnegative 
sparse coding problem. 
\begin{align}
\boldsymbol{w}_{j} & \coloneqq\phi_{j}\Big(-\mathbf{B}_{jj}^{\mathsf{T}}\sum_{k=1}^{j-1}\mathbf{B}_{jk}^{\mathsf{T}}\boldsymbol{w}_{k}\Big)\quad \forall j=1,\dotsc,l
\label{eq:dense}
\\[-0.5em]
 & \approx\underset{\{\boldsymbol{w}_{j}\}}{\arg\min}\sum_{j=1}^{l}\Big\Vert \mathbf{B}_{jj}\boldsymbol{w}_{j}+\sum_{k=1}^{j-1}\mathbf{B}_{jk}^{\mathsf{T}}\boldsymbol{w}_{j}\Big\Vert _{F}^{2}+\Phi_{j}(\boldsymbol{w}_{j})\nonumber
\end{align}
In comparison to Eq.~\ref{eq:chain}, additional parameters introduce skip connections 
between layers so that the activations $\boldsymbol{w}_j$ of layer $j$ now depend on those of 
all previous layers $k<j$.

These connections are similar to the identity mappings in residual networks~\cite{he2016deep}, which introduce 
dependence between the activations of pairs of layers for even $j\in[1,l-1]$:
\begin{equation}
\hspace{-0.5em}\boldsymbol{w}_{j} \coloneqq\phi_{j}(\mathbf{B}_{j}^{\mathsf{T}}\boldsymbol{w}_{j-1}),\,
\boldsymbol{w}_{j+1} \coloneqq\phi_{j+1}(\boldsymbol{w}_{j-1}{+}\mathbf{B}_{j+1}^{\mathsf{T}}\boldsymbol{w}_{j})
\hspace{-0.2em}\label{eq:residual_act}
\end{equation}
In comparison to chain networks, no additional parameters are
required; the only difference is the 
addition of $\boldsymbol{w}_{j-1}$ in the argument of $\phi_{j+1}$.
As a special case of Eq.~\ref{eq:dense}, we interpret the activations in Eq.~\ref{eq:residual_act} as approximate  
solutions to the optimization problem:
\begin{gather}
\underset{\{\boldsymbol{w}_{j}\}}{\arg\min} \left\lVert \boldsymbol{x}-\mathbf{B}_{1}\boldsymbol{w}_{1}\right\rVert _{2}^{2} + \sum_{j=1}^{l}\Phi_{j}(\boldsymbol{w}_{j}) \label{eq:residual_opt} \\[-0.5em]{+}\hspace{-0.1em}\sum_{\mathrm{even}\,j}\hspace{-0.1em}\left\lVert \boldsymbol{w}_{j}-\mathbf{B}_{j}^\mathsf{T}\boldsymbol{w}_{j-1}\right\rVert _{2}^{2} 
{+}\left\lVert \boldsymbol{w}_{j+1}-\boldsymbol{w}_{j-1}-\mathbf{B}_{j+1}^{\mathsf{T}}\boldsymbol{w}_{j}\right\rVert _{2}^{2}
\nonumber
\end{gather}
This results in the induced dictionary structure 
of Eq.~\ref{eq:dense_structure} with $\mathbf{B}_{jj}=\mathbf{I}$
for $j>1$, $\mathbf{B}_{jk}=\mathbf{0}$ for $j>k+1$, $\mathbf{B}_{jk}=\mathbf{0}$ for $j>k$ with odd $k$, and 
$\mathbf{B}_{jk}=\mathbf{I}$ for $j>k$ with even $k$.

Building upon the empirical successes of residual networks, densely connected convolution networks~\cite{huang2017densely} incorporate 
skip connections between earlier layers as well. This is shown in Eq.~\ref{eq:densenet_act} 
where the transformation $\mathbf{B}_j$ of concatenated variables $[\boldsymbol{w}_{k}]_k$ for $k=1,\dotsc,j-1$ 
is equivalently 
written as the summation of smaller transformations $\mathbf{B}_{jk}$.
\begin{equation}
\boldsymbol{w}_{j} \coloneqq\phi_{j}\Big(\mathbf{B}_{j}^{\mathsf{T}}\left[\boldsymbol{w}_{k}\right]_{k=1}^{j-1}\Big)=\phi_{j}\Big(\sum_{k=1}^{j-1}\mathbf{B}_{jk}^{\mathsf{T}}\boldsymbol{w}_{k}\Big)
\label{eq:densenet_act}
\end{equation}
These activations again provide approximate solutions to the problem in Eq.~\ref{eq:dense} with 
the induced dictionary structure of Eq.~\ref{eq:dense_structure} where 
$\mathbf{B}_{jj}=\mathbf{I}$ for $j>1$ and the lower blocks $\mathbf{B}_{jk}$ for $j>k$ are all 
filled with learned parameters. 

Skip connections enable effective learning 
in much deeper networks than chain-structured alternatives. While originally motivated from the perspective of 
making optimization easier~\cite{he2016deep}, adding more connections between layers was also shown to 
improve generalization performance~\cite{huang2017densely}.
As compared in Fig.~\ref{fig:gram}, denser skip connections induce dictionary structures with denser Gram 
matrices allowing for lower mutual coherence. This suggests that architectures' capacities for low 
validation error can be quantified and compared based on their capacities for 
inducing dictionaries with low minimum mutual coherence. 

\section{The Deep Frame Potential}

\begin{figure}
\centering

\hspace*{\fill}
\subfloat[Convolutional Dictionary]{
\includegraphics[width=0.4\columnwidth]{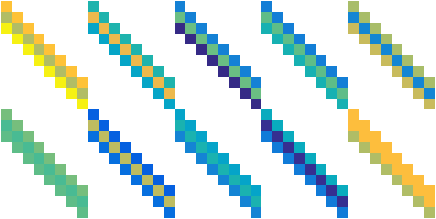} 
}
\hspace*{\fill}
\hspace*{\fill}
\subfloat[Permuted Dictionary]{
\includegraphics[width=0.4\columnwidth]{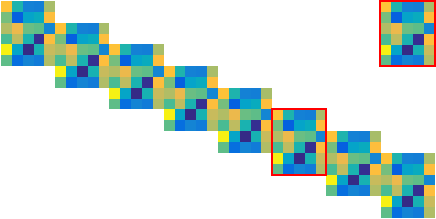}
}
\hspace*{\fill}

\hspace*{\fill}
\subfloat[Convolutional Gram Matrix]{
\includegraphics[width=0.4\columnwidth]{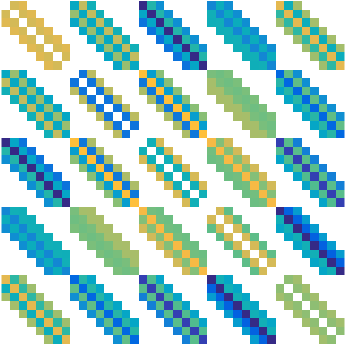}
}
\hspace*{\fill}
\hspace*{\fill}
\subfloat[Permuted Gram Matrix]{
\includegraphics[width=0.4\columnwidth]{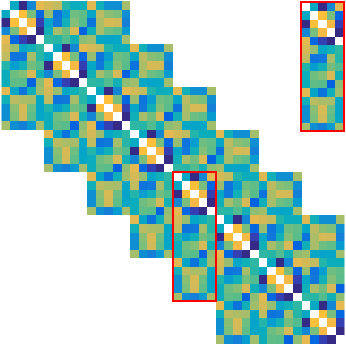}
}
\hspace*{\fill}

\caption{
\small A visualization of a one-dimensional convolutional dictionary with two input 
channels, five output channels, and a filter size of three. (a) The filters are repeated over
eight spatial dimensions resulting in a (b) block-Toeplitz structure that is
revealed through row and column permutations. (c) The corresponding gram matrix can be 
efficiently computed by (d) repeating local filter interactions. 
}
\label{fig:conv}
\end{figure}

We propose to use lower bounds on the mutual coherence of induced 
structured dictionaries for the data-independent comparison of architecture capacities. 
Note that while one-sided coherence is better suited to nonnegativity constraints, it has the 
same lower bound~\cite{bruckstein2008uniqueness}.
Directly optimizing mutual coherence from Eq.~\ref{eq:mutual_coherence}
is difficult due to its piecewise structure. Instead, we consider a  
tight lower by replacing the maximum off-diagonal element of the
Gram matrix $\mathbf{G}$ with the mean.
This gives the averaged frame potential $F^2(\mathbf{B})$,
a strongly-convex function that can be optimized more effectively~\cite{benedetto2003finite}:
\begin{equation}
F^2(\mathbf{B})=N^{-1}(\mathbf{G})\left(\left\lVert \mathbf{G}\right\rVert _{F}^{2}-\mathrm{Tr}\,\mathbf{G}\right) \leq \mu^{2}(\mathbf{B})
\label{eq:welch_bound}
\end{equation}
Here, $N(\mathbf{G})$ is the number of nonzero off-diagonal elements in the Gram matrix 
and $\mathrm{Tr}\,\mathbf{G}$ equals the total number of dictionary atoms.
Equality is met in the case of 
equiangular tight frames when the normalized inner products between all dictionary atoms are 
equivalent~\cite{kovavcevic2008introduction}.
Due to the block-sparse structure of the induced dictionaries from Eq.~\ref{eq:dense_structure}, we  evaluate the frame potential in terms of local blocks $\mathbf{G}_{jj'}\in\mathbb{R}^{k_j\times k_{j'}}$ that are nonzero
only if layer $j$ is connected to layer $j'$. In the case of convolutional layers 
with localized spatial support, there is also a repeated implicit structure of nonzero elements as visualized in 
Fig.~\ref{fig:conv}.

To compute the Gram matrix, we first need to normalize the global induced dictionary $\mathbf{B}$
from Eq.~\ref{eq:dense_structure}.
By stacking the column magnitudes of layer $j$ as the elements in the diagonal matrix 
$\mathbf{C}_{j}=\mathrm{diag}(\boldsymbol{c}_{j})\in\mathbb{R}^{k_{j}\times k_{j}}$, the normalized 
parameters can be represented as $\tilde{\mathbf{B}}_{ij} = \mathbf{B}_{ij}\mathbf{C}_{j}^{-1}$. 
Similarly, the squared norms of the full set of columns in the global dictionary $\mathbf{B}$ are  
$\mathbf{N}_{j}^{2}=\sum_{i=j}^{l}\mathbf{C}_{ij}^{2}$.
The full normalized dictionary can then be found as  $\tilde{\mathbf{B}}=\mathbf{B}\mathbf{N}^{-1}$
where the matrix $\mathbf{N}$ is block diagonal with $\mathbf{N}_{j}$ as its blocks.
The blocks of the Gram matrix 
$\mathbf{G}=\tilde{\mathbf{B}}^\mathsf{T}\tilde{\mathbf{B}}$ are then given as:
\begin{equation}
\mathbf{G}_{jj'}=\sum_{i=j'}^{l}\mathbf{N}_{j}^{-1}\mathbf{B}_{ij}^{\mathsf{T}}\mathbf{B}_{ij'}\mathbf{N}_{j'}^{-1}
\label{eq:G_block}
\end{equation}
For chain networks, $\mathbf{G}_{jj'}\neq\mathbf{0}$ only when $j'=j+1$, which represents the 
connections between adjacent layers. In this case, the blocks can be simplified as:
\begin{align}
\mathbf{G}_{jj} & =(\mathbf{C}_{j}^{2}+\mathbf{I})^{-\frac{1}{2}}({\mathbf{B}}_{j}^{\mathsf{T}}{\mathbf{B}}_{j}+\mathbf{I})(\mathbf{C}_{j}^{2}+\mathbf{I})^{-\frac{1}{2}}\\
\mathbf{G}_{j(j+1)} & =-(\mathbf{C}_{j}^{2}+\mathbf{I})^{-\frac{1}{2}}{\mathbf{B}}_{j+1}(\mathbf{C}_{j+1}^{2}+\mathbf{I})^{-\frac{1}{2}}\label{eq:G_block_chain}\\
\mathbf{G}_{ll}&={\mathbf{B}}_{l}^{\mathsf{T}}{\mathbf{B}}_{l}
\end{align}

Because the diagonal is removed in the deep frame potential computation, the contribution of $\mathbf{G}_{jj}$ 
is simply a rescaled version of the local frame potential of layer $j$. The contribution of $\mathbf{G}_{j(j+1)}$,
on the other hand, can essentially be interpreted as rescaled $\ell_2$ weight decay where rows are
weighted more heavily if the corresponding columns of the previous layer's parameters have higher magnitudes.
Furthermore, since the global frame potential is averaged over the total number of nonzero elements in $\mathbf{G}$,
if a layer has more parameters, then it will be given more weight in this computation. 
For more general networks with skip connections, however, the summation from Eq.~\ref{eq:G_block} 
has additional terms that introduce 
more complicated interactions. In these cases, it cannot be evaluated from local properties of  layers.

Essentially, the deep frame potential summarizes the structural properties of
the global dictionary $\mathbf{B}$
induced by a deep network architecture by balancing interactions within each individual layer through
local coherence properties and between connecting layers.

\subsection{Theoretical Lower Bound for Chain Networks}

While the deep frame potential is a function of parameter values, its minimum value is determined 
only by the architecture-induced dictionary structure. Furthermore, we know that
it must be bounded by a nonzero constant for overcomplete dictionaries. 
In this section, we 
derive this lower bound for the special case of fully-connected chain networks
and provide intuition for why skip connections increase the capacity for low mutual coherence.

First, observe that a lower bound for the Frobenius norm of $\mathbf{G}_{j(j+1)}$ from 
Eq.~\ref{eq:G_block_chain} cannot be readily attained because the rows and columns are
rescaled independently. This means that a lower bound for the norm 
of $\mathbf{G}$ must be found by jointly considering the entire matrix structure, 
not simply through
the summation of its components. To accomplish this, we instead consider the matrix 
$\mathbf{H}=\tilde{\mathbf{B}}\tilde{\mathbf{B}}^{\mathsf{T}}$, which is full rank and has the same norm as 
$\mathbf{G}$:
\begin{equation}
\lVert \mathbf{G} \rVert _{F}^{2}=\lVert \mathbf{H} \rVert _{F}^{2}=\sum_{j=1}^{l}\left\lVert \mathbf{H}_{jj}\right\rVert _{F}^{2}+2\sum_{j=1}^{l-1}\left\lVert \mathbf{H}_{j(j+1)}\right\rVert _{F}^{2}
\label{eq:H_norm}
\end{equation}
We can then express the individual blocks of $\mathbf{H}$ as:
\begin{align}
\mathbf{H}_{11} & ={\mathbf{B}}_{1}(\mathbf{C}_{1}^{2}+\mathbf{I}_{k_{1}})^{-1}{\mathbf{B}}_{1}^{\mathsf{T}}\\
\mathbf{H}_{jj} & ={\mathbf{B}}_{j}(\mathbf{C}_{j}^{2}+\mathbf{I})^{-1}{\mathbf{B}}_{j}^{\mathsf{T}}+(\mathbf{C}_{j-1}^{2}+\mathbf{I})^{-1}\\
\mathbf{H}_{j(j+1)} & =-{\mathbf{B}}_{j}(\mathbf{C}_{j}^{2}+\mathbf{I})^{-1}
\label{eq:Hjj}
\end{align}
In contrast to $\mathbf{G}_{j(j+1)}$ in Eq.~\ref{eq:G_block_chain}, only the columns of 
$\mathbf{H}_{j(j+1)}$ in Eq.~\ref{eq:Hjj} are rescaled.
Since $\tilde{\mathbf{B}}_j$ has normalized columns, its norm can be exactly expressed
as:
\begin{equation}
\left\Vert \mathbf{H}_{j(j+1)}\right\rVert _{F}^{2} =\sum_{n=1}^{k_{j}}\Bigg(\frac{c_{jn}}{c_{jn}^{2}+1}\Bigg)^{2}
\label{eq:H_jk}
\end{equation}

For the other blocks, we find lower bounds for their norms through the same technique used in deriving
the Welch bound, which expresses the minimum mutual coherence for unstructured overcomplete  
dictionaries~\cite{welch1974lower}. 
Specifically, we apply the Cauchy-Schwarz inequality giving 
$\left\lVert \mathbf{A}\right\rVert _{F}^{2} \geq r^{-1}(\mathrm{Tr}\,\mathbf{A})^2$ for positive-semidefinite
matrices $\mathbf{A}$ with rank $r$. Since the rank of $\mathbf{H}_{jj}$ is at most $k_{j-1}$, we can lower bound the norms of the individual blocks as:
\begin{align}
\left\lVert \mathbf{H}_{11}\right\rVert _{F}^{2} & \geq\frac{1}{k_{0}}\Bigg(\sum_{n=1}^{k_{1}}\frac{c_{1n}^{2}}{c_{1n}^{2}+1}\Bigg)^{2}\\
\left\lVert \mathbf{H}_{jj}\right\rVert _{F}^{2} & \geq\frac{1}{k_{j-1}}\Bigg(\sum_{n=1}^{k_{j}}\frac{c_{jn}^{2}}{c_{jn}^{2}+1}+\sum_{p=1}^{k_{j-1}}\frac{1}{c_{(j-1)p}^{2}+1}\Bigg)^{2}\nonumber
\label{eq:Hjj_bound}
\end{align}

In this case of dense shallow dictionaries, the Welch bound depends only on the data dimensionality
and the number of dictionary atoms. However, due to the structure
of the architecture-induced dictionaries, the lower bound of the deep frame potential depends on the data
dimensionality, the number of layers, the number of units in each layer, the connectivity between layers,
and the relative magnitudes between layers. Skip connections increase the number of nonzero elements
in the Gram matrix over which to average and also enable off-diagonal blocks to have lower norms.

\subsection{Model Selection}

For more general architectures that lack a simple theoretical lower bound, we instead propose 
bounding the mutual coherence of the architecture-induced dictionary through 
empirical minimization of the deep frame potential $F^2(\mathbf{B})$ from Eq.~\ref{eq:welch_bound}. 
Frame potential minimization has been used effectively to construct finite normalized tight 
frames due to the lack of suboptimal local minima, which allows for effective optimization
using gradient descent~\cite{benedetto2003finite}. We propose using the minimum deep
frame potential of an architecture--which is independent of data and individual
parameter instantiations--as a means to compare different architectures. In practice, model 
selection is performed by choosing the candidate architecture with the lowest minimum frame potential
subject to desired modeling constraints such as limiting the total number of parameters.  

\section{Experimental Results}

In this section, we demonstrate correlation between the minimum deep frame potential and 
validation error on the CIFAR-10 dataset~\cite{krizhevsky2009learning} across a wide variety of 
fully-connected, convolutional, chain, residual, and densely connected network architectures. Furthermore, 
we show that networks with skip connections can have lower deep frame potentials with 
fewer learnable parameters, which is predictive of the parameter efficiency of trained networks. 

\begin{figure}
\centering
\subfloat[Without Regularization]{
\includegraphics[width=0.48\columnwidth]{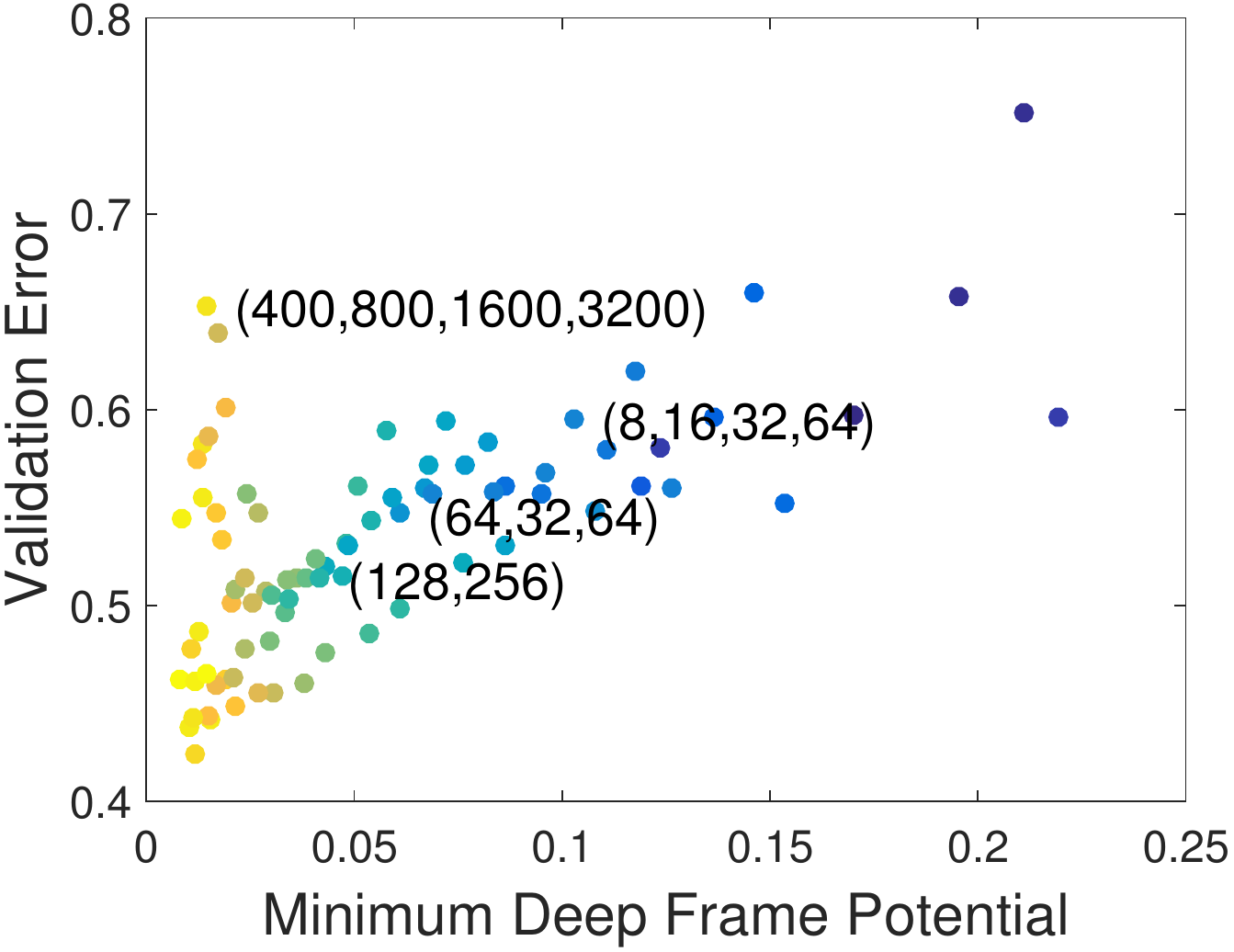}
}
\subfloat[With Regularization]{
\includegraphics[width=0.48\columnwidth]{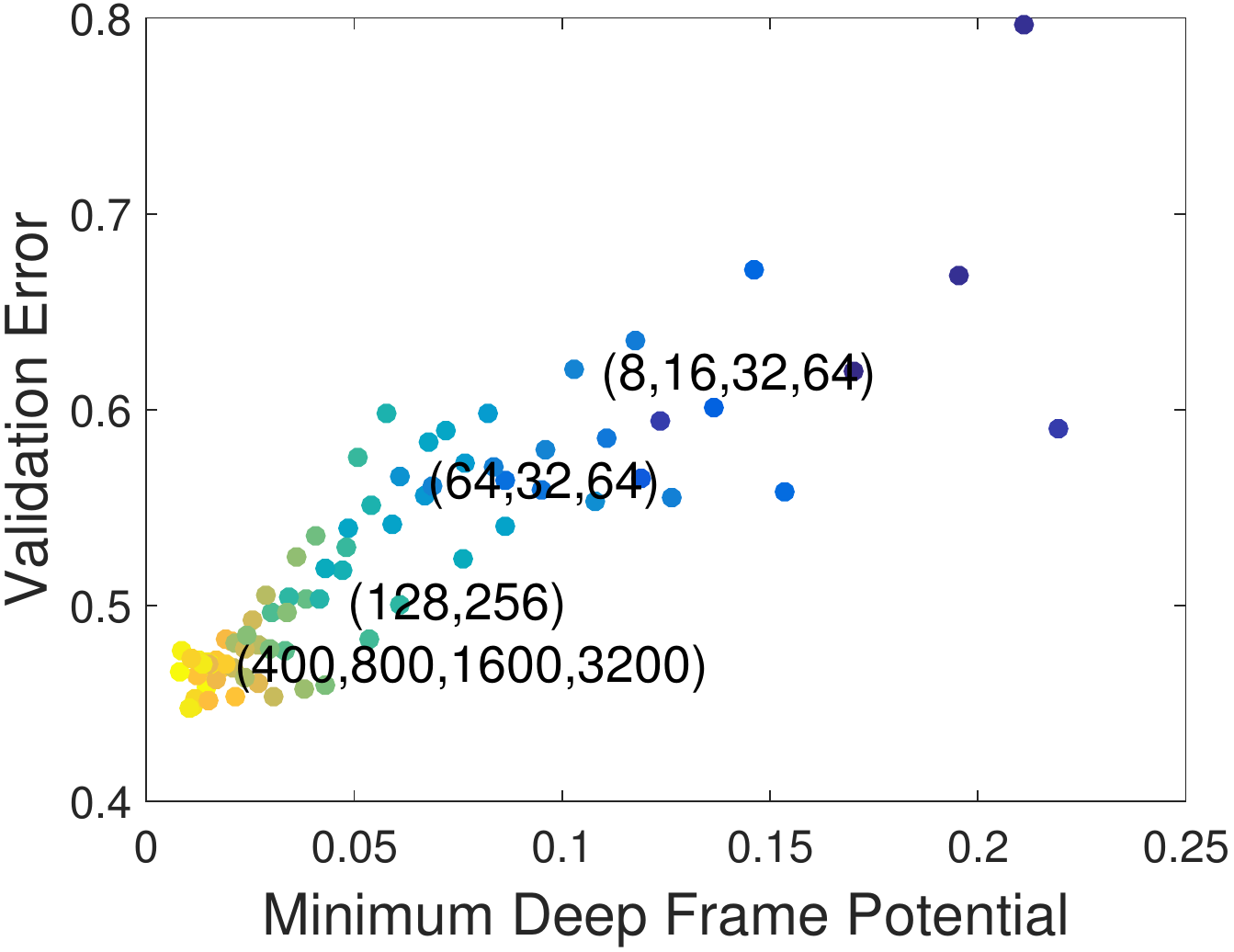}
}
\caption{
\small A comparison of fully connected 
deep network architectures with varying depths and widths. Warmer colors indicate models with more total parameters. 
(a) Some very large networks cannot be trained effectively resulting in unusually high validation errors. 
(b) This can be remedied through deep frame potential regularization, resulting in high correlation between 
minimum frame potential and validation error.
}
\label{fig:regularization}
\end{figure}

In Fig.~\ref{fig:regularization}, we visualize a scatter plot of trained  
fully-connected networks with between three and five layers and between 16 and 4096 units 
in each layer. The corresponding architectures are shown as a list of units per layer
for a few representative examples. 
The minimum frame potential of each architecture is compared against its validation 
error after training, and the total parameter count is indicated by color.
In Fig.~\ref{fig:regularization}a, some networks with many parameters have
unusually high error due to the difficulty in training very large fully-connected networks. 
In Fig.~\ref{fig:regularization}b, the addition of a deep frame
potential regularization term overcomes some of these optimization difficulties for improved 
parameter efficiency. This results in high
correlation between minimum frame potential and validation error. Furthermore, it emphasizes the  
diminishing returns of increasing the size of fully-connected chain networks; after a certain point, adding more 
parameters does little to reduce both validation error and minimum frame potential.

\begin{figure}
\centering

\subfloat[Validation Error]{
\includegraphics[width=0.48\columnwidth]{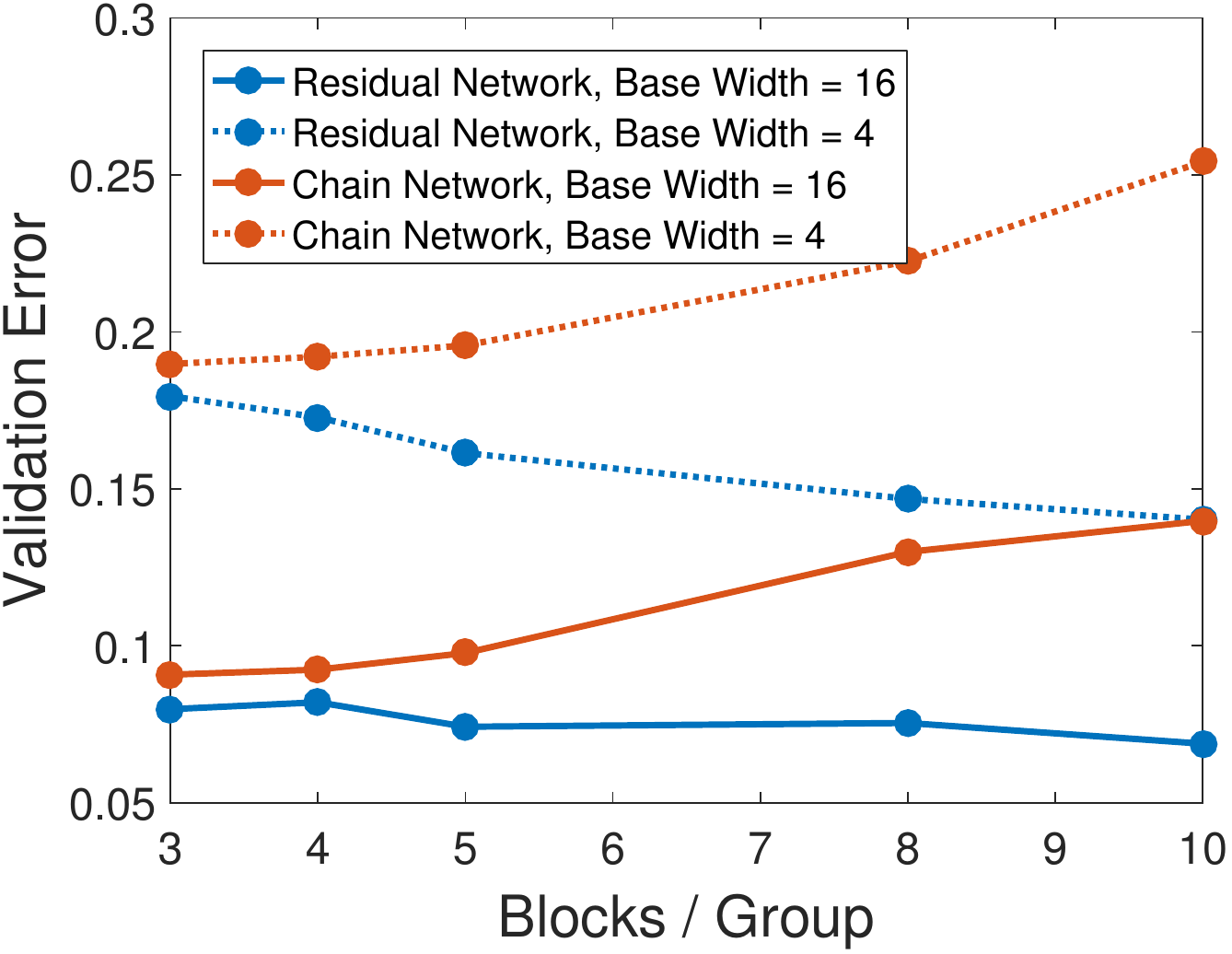}
}
\subfloat[Minimum Deep Frame Potential]{
\includegraphics[width=0.48\columnwidth]{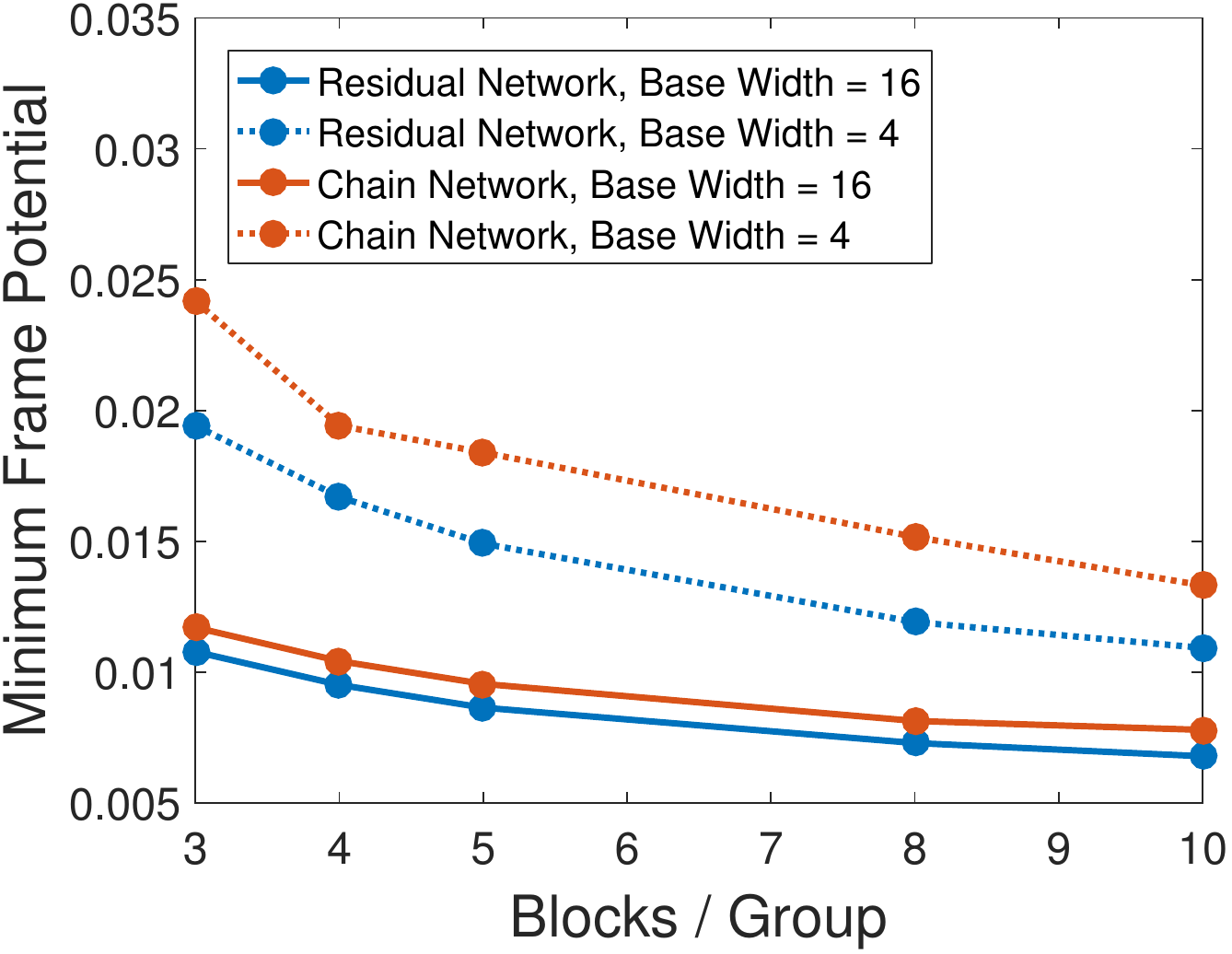}
}
\caption{
\small The effect of increasing depth in chain and residual networks.
Validation error is compared against layer count for two different network widths. (a) In comparison to 
chain networks, even very deep residual networks can be trained effectively resulting in decreasing
validation error. (b) Despite having the same number of total parameters, residual connections also 
induce dictionary structures with lower minimum deep frame potentials.
}
\label{fig:resnet_layers}
\end{figure}

To evaluate the effects of residual connections~\cite{he2016deep}, we adapt the 
simplified CIFAR-10 ResNet architecture from~\cite{wu2016tensorpack}, which consists of a single convolutional 
layer followed 
by three groups of residual blocks with activations given in Eq.~\ref{eq:residual_act}. Before the second and
third groups, the number of filters is increased by a factor of two and the spatial resolution is 
decreased by half through average pooling. 
To compare networks with different sizes, we modify their depths by changing the number of residual blocks 
in each group from between 2 and 10 and their widths by changing the base number of filters from between 
4 and 32.
For our experiments with densely connected skip connections~\cite{huang2017densely}, we adapt the
simplified CIFAR-10 DenseNet architecture from~\cite{li2018densepack}. Like with residual networks, 
it consists of a convolutional layer followed by three groups of the activations 
from Eq.~\ref{eq:densenet_act}
with decreasing 
spatial resolutions and increasing numbers of filters. Within each group, a dense block is 
the concatenation of smaller convolutions that take all previous outputs as inputs with filter numbers
equal to a fixed growth rate. Network depth and width are modified by respectively increasing both 
the number of 
layers per group and the base growth rate from between 2 and 12. Batch normalization~\cite{ioffe2015batch} 
was also used in all convolutional networks.

In Fig.~\ref{fig:resnet_layers}, we compare the validation errors and minimum frame potentials of
residual networks and comparable chain networks with residual connections removed. 
In Fig.~\ref{fig:resnet_layers}a, the validation error of chain networks increases for  
deeper networks while that of residual networks is lower and consistently decreases. 
This emphasizes the difficulty in training very deep chain networks. 
In Fig.~\ref{fig:resnet_layers}b, we show that residual connections enable lower minimum frame 
potentials following a similar trend with respect to
increasing model size, again demonstrating correlation between validation error and minimum frame potential. 

\begin{figure}
\centering

\subfloat[Validation Error]{
\includegraphics[width=0.48\columnwidth]{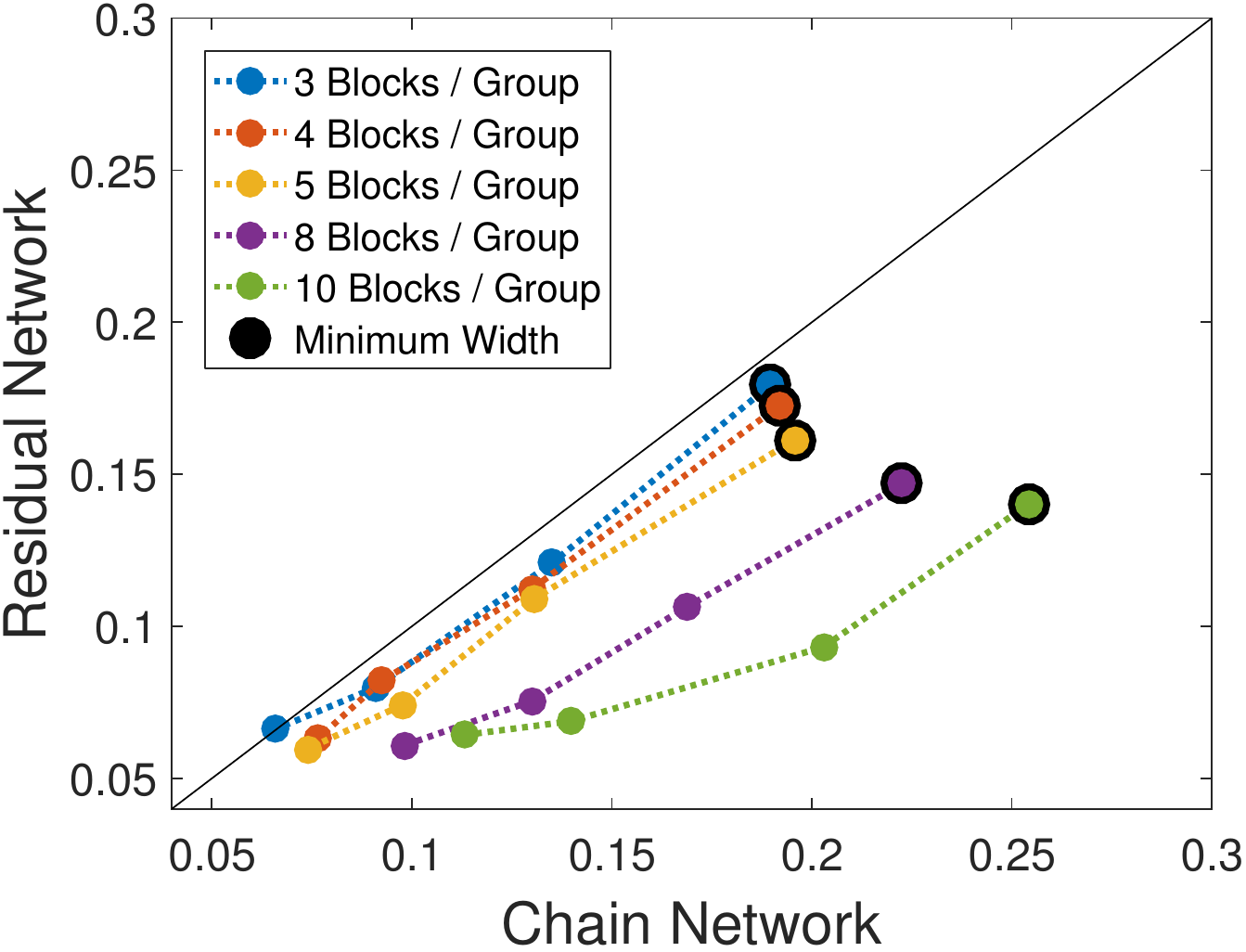}
}
\subfloat[Minimum Deep Frame Potential]{
\includegraphics[width=0.48\columnwidth]{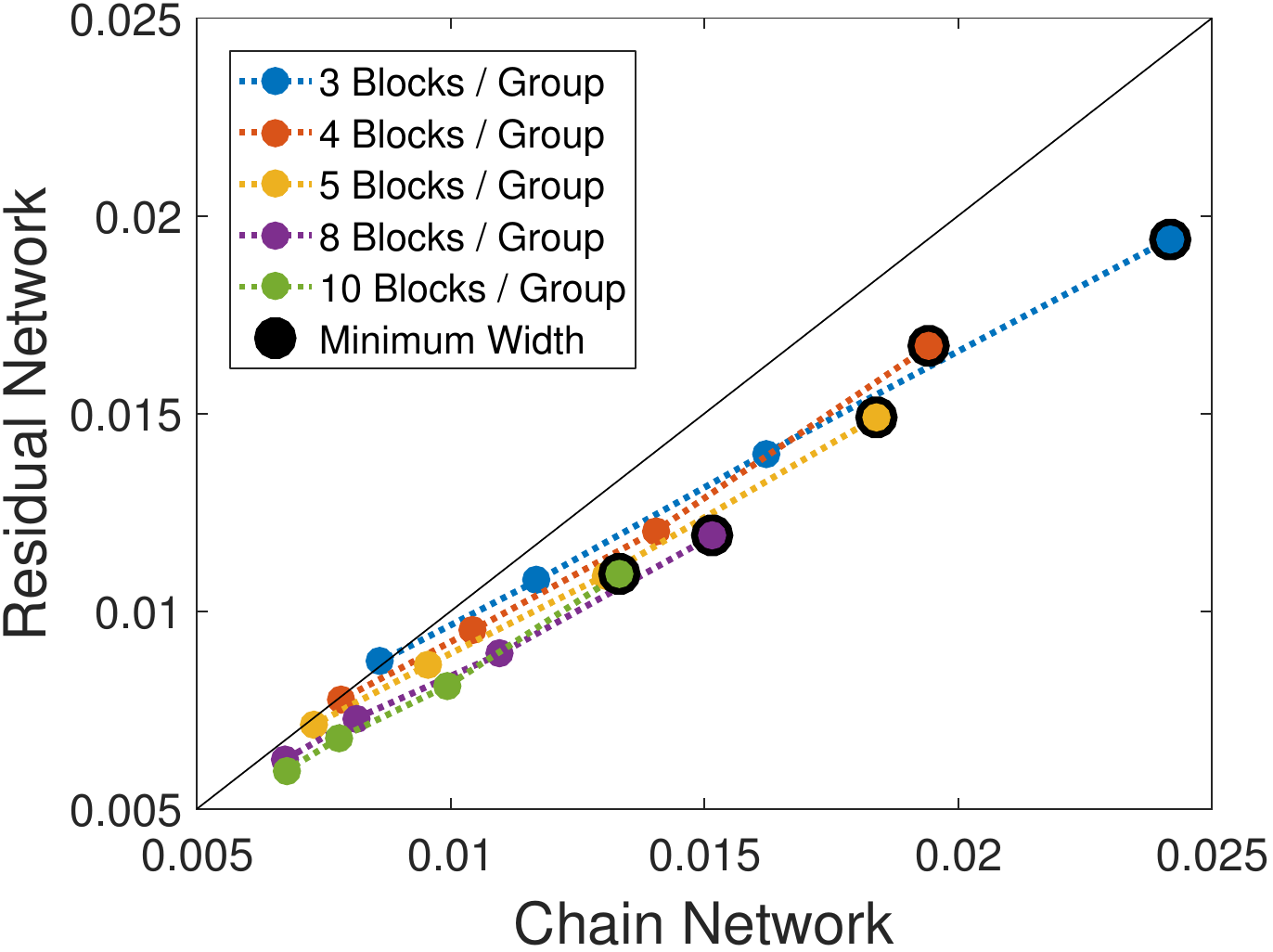}
}

\caption{
\small A comparison of (a) validation error and (b) minimum frame potential between  
residual networks and chain networks. Colors indicate different depths 
and datapoints are connected in order of increasing widths of 4, 8, 16, or 32 filters. 
Skip connections result in 
reduced error correlating with frame potential with dense networks showing 
superior efficiency with increasing depth.
}
\label{fig:scatter}
\end{figure}

In Fig.~\ref{fig:scatter}, we compare chain networks and residual 
networks with exactly the same number of parameters, where color 
indicates the number of residual blocks per group and connected data points 
have the same depths but different widths.
The addition
of skip connections reduces both validation error and minimum frame potential, as visualized 
by consistent placement below the diagonal line indicating lower values for residual networks than 
comparable chain networks.
This effect becomes even more pronounced with increasing depths and widths. 

\begin{figure}
\centering

\subfloat[Chain Validation Error]{
\includegraphics[width=0.48\columnwidth]{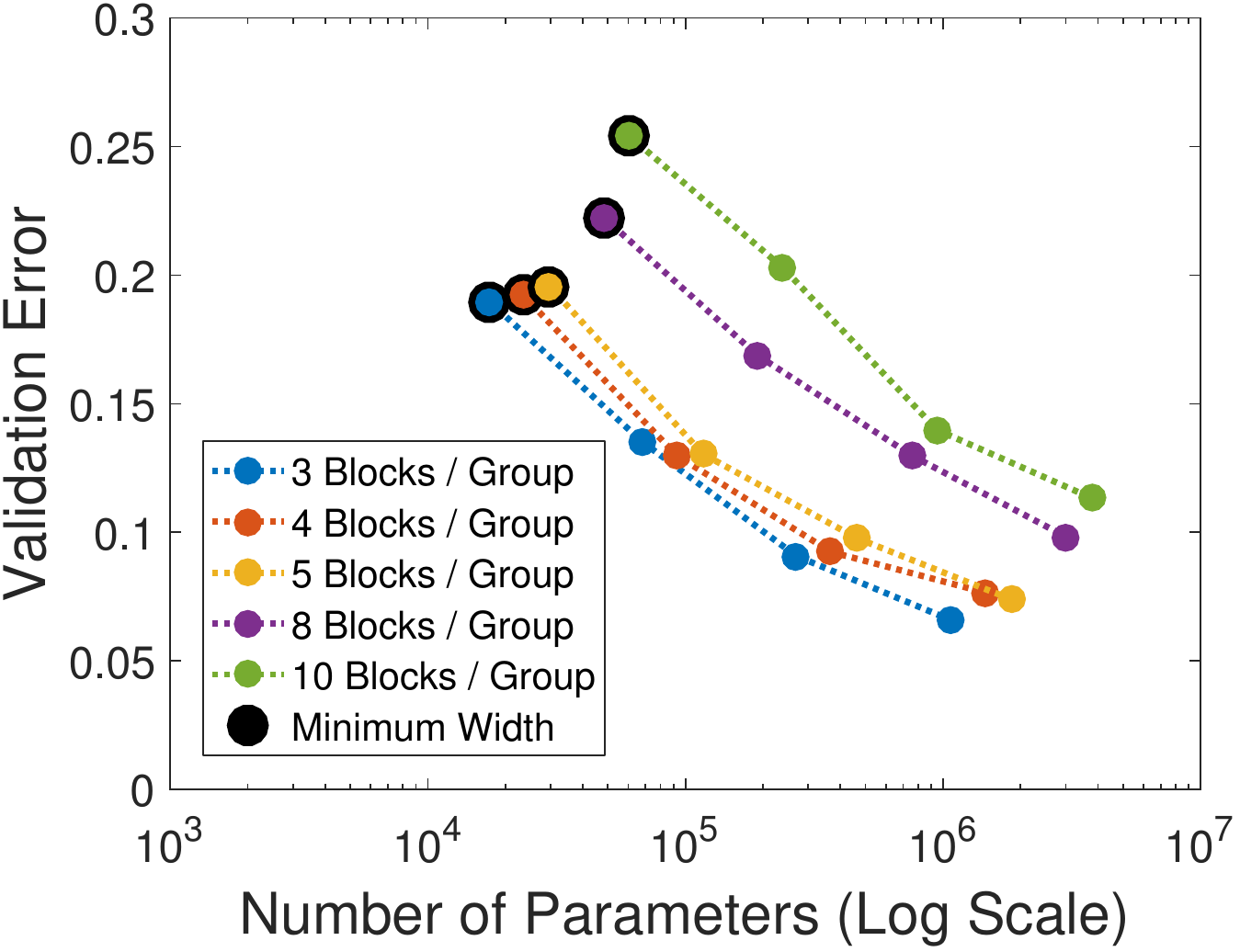}
}
\subfloat[Chain Frame Potential]{
\includegraphics[width=0.48\columnwidth]{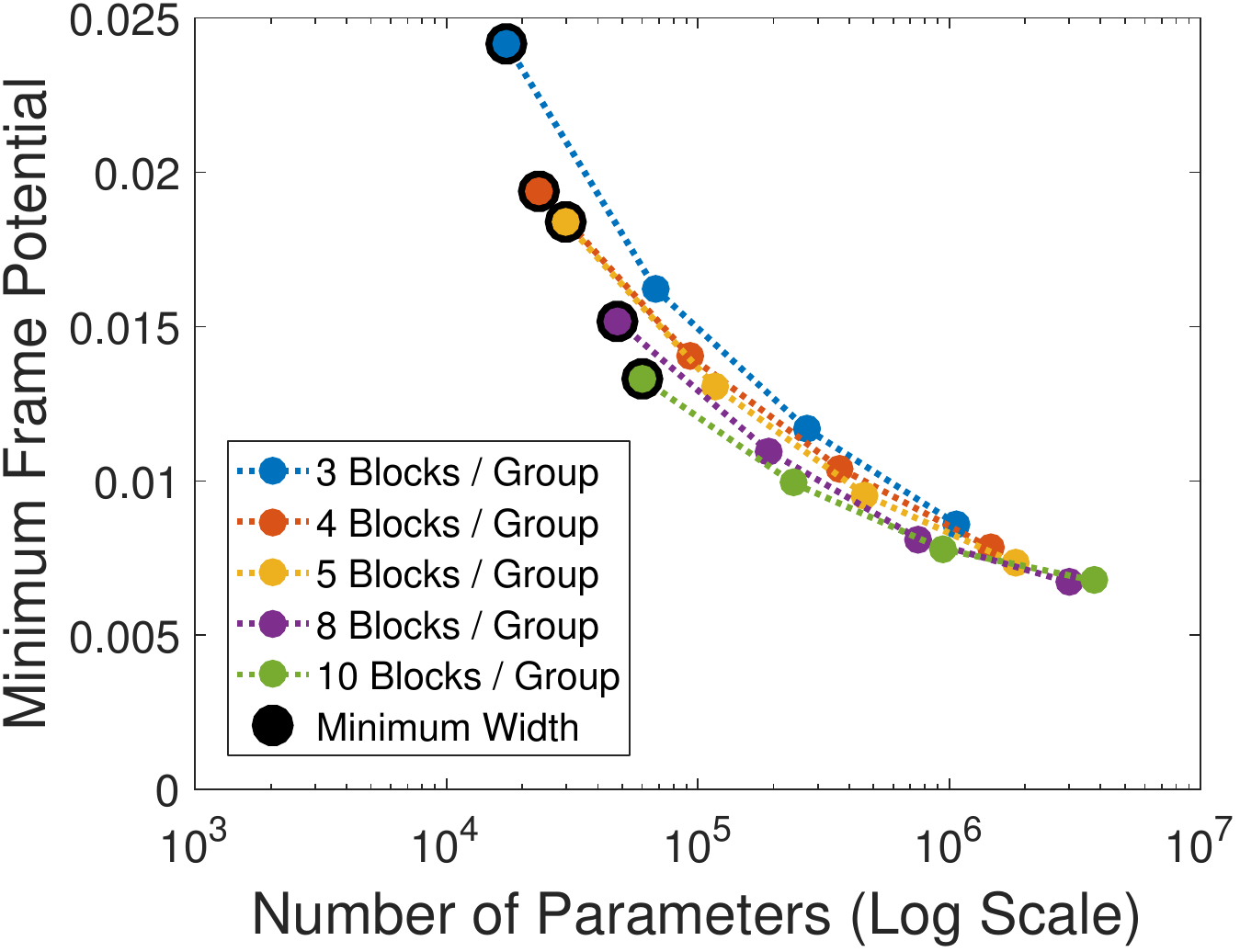}
}

\subfloat[ResNet Validation Error]{
\includegraphics[width=0.48\columnwidth]{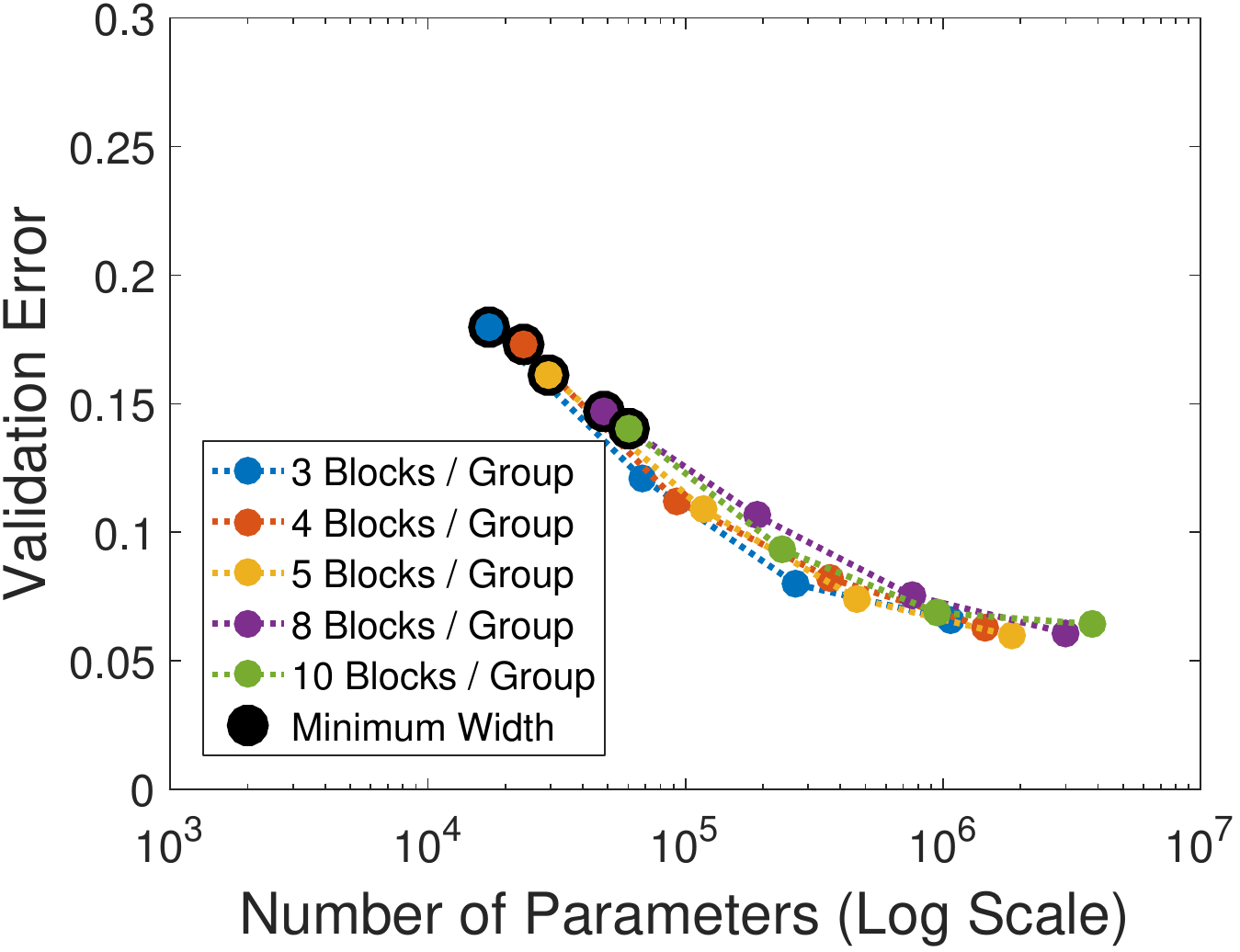}
}
\subfloat[ResNet Frame Potential]{
\includegraphics[width=0.48\columnwidth]{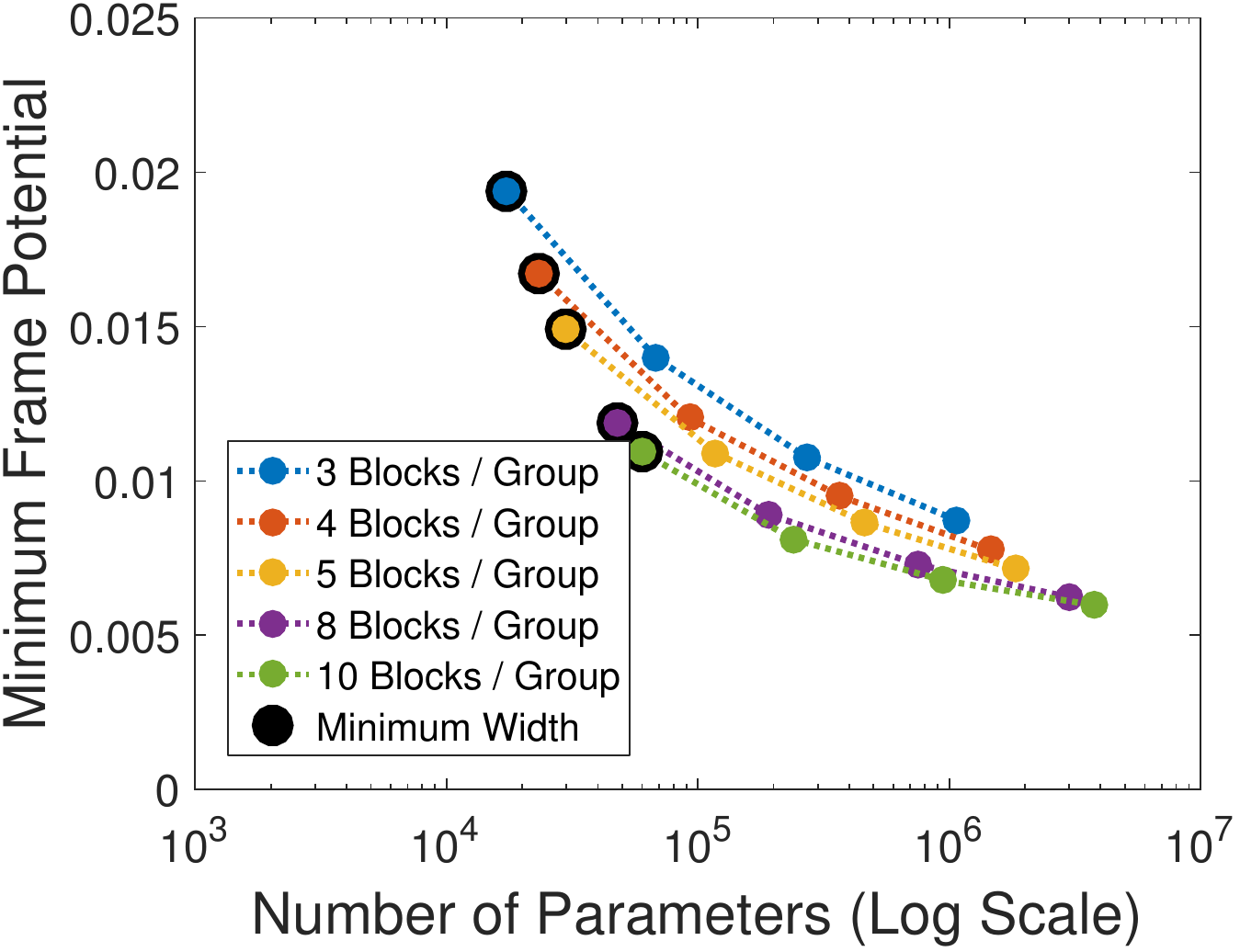}
}

\subfloat[DenseNet Validation Error]{
\includegraphics[width=0.48\columnwidth]{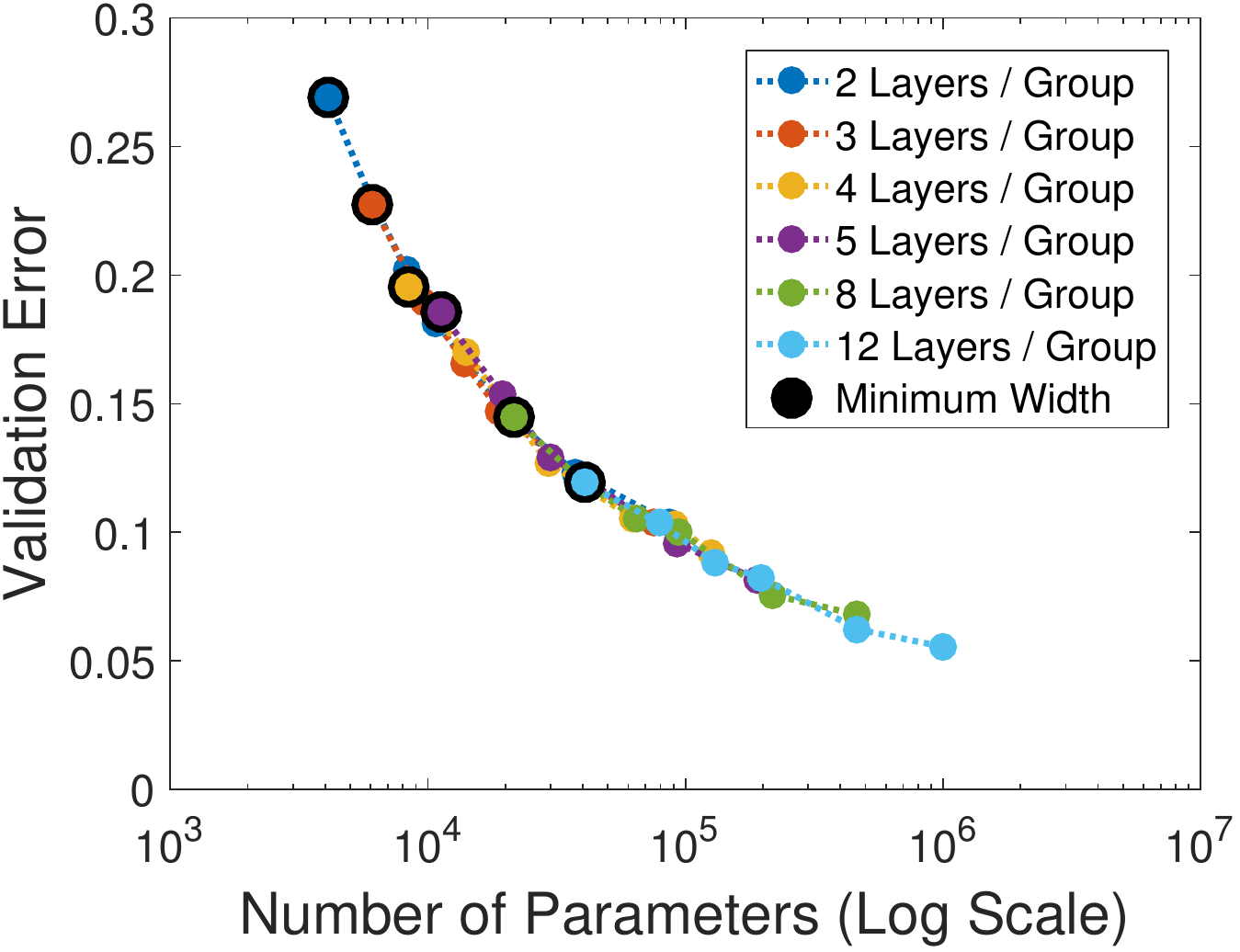}
}
\subfloat[DenseNet Frame Potential]{
\includegraphics[width=0.48\columnwidth]{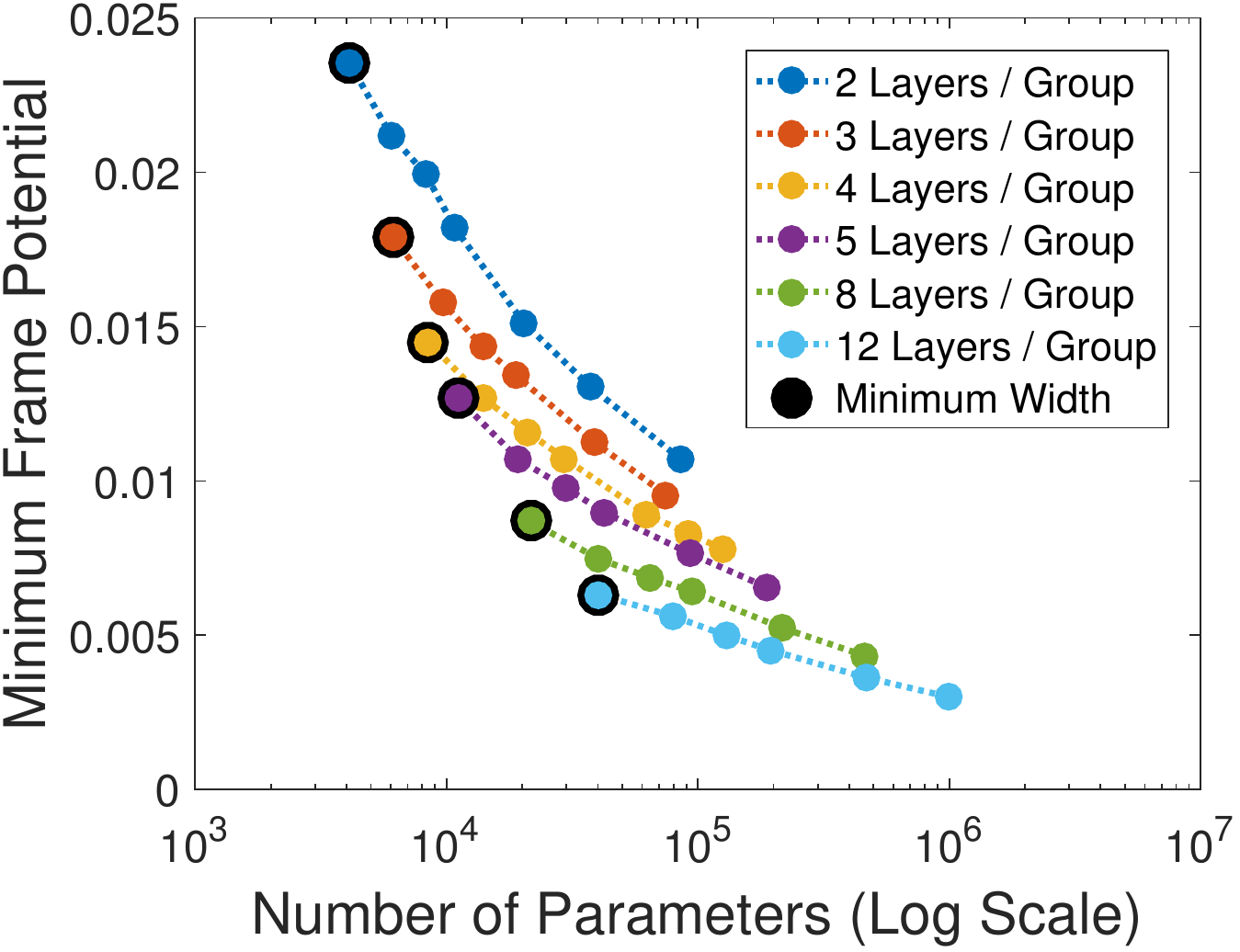}
}
\caption{
\small A demonstration of the improved scalability of networks with skip connections, where line
colors indicate 
different depths and data points are connected showing increasing widths. (a) Chain networks with greater 
depths have
increasingly worse parameter efficiency in comparison to (c) the corresponding networks with residual
connections and (e) densely connected networks with similar size, of which performance scales 
efficiently with parameter count. This
could potentially be attributed to correlated efficiency in reducing frame potential with fewer 
parameters, which saturates much faster with  (b) chain networks than (d) residual networks or 
(f) densely connected networks.
}
\label{fig:resnet_width}
\end{figure}

In Fig.~\ref{fig:resnet_width}, we compare the parameter efficiency of chain networks, 
residual networks, and densely connected networks of different depths and widths. 
We visualize both validation error and 
minimum frame potential as functions of the number of parameters, 
demonstrating the improved scalability of 
networks with skip connections.
While chain networks demonstrate increasingly poor 
parameter efficiency with respect to increasing depth in 
Fig.~\ref{fig:resnet_width}a, the skip connections of ResNets and DenseNets 
allow for further reducing error with larger network sizes in 
Figs.~\ref{fig:resnet_width}c,e. Considering all network families together as in Fig.~\ref{fig:gram}d, 
we see that denser connections also allow for lower validation error with comparable numbers of parameters.
This trend is mirrored in the minimum frame potentials of Figs.~\ref{fig:resnet_width}b,d,f
which are shown together in Fig.~\ref{fig:gram}e. Despite some fine variations in behavior across
different families of architectures, minimum deep frame potential is correlated with validation error across
network sizes and effectively predicts the increased generalization capacity provided by skip connections.

\section{Conclusion}

In this paper, we proposed a technique for comparing deep network architectures by approximately
quantifying their implicit capacity for effective data representations.
Based upon 
theoretical 
connections between sparse approximation and deep neural networks, we demonstrated how architectural
hyper-parameters such as 
depth, width, and skip connections induce different structural properties
of the dictionaries in corresponding sparse coding problems. 
We compared these dictionary structures through lower
bounds on their mutual coherence, which is theoretically tied to their capacity for uniquely
and robustly representing data via sparse approximation. A theoretical lower bound was derived 
for chain networks and the deep frame potential was proposed as an empirical optimization objective
for constructing bounds for more complicated architectures. 

Experimentally, we observed
a correlation between minimum deep frame potential and validation error across different families of 
modern architectures with skip connections, including residual networks and densely connected convolutional networks. 
This suggests a
promising direction
for future research towards the theoretical analysis and practical construction of deep network
architectures derived from connections between deep learning and sparse coding. 
{\bf Acknowledgments:} This work was supported by the CMU Argo AI Center for Autonomous Vehicle Research
and by the National Science Foundation under Grant No.1925281.

\clearpage
{\small
\bibliographystyle{ieee_fullname}
\bibliography{refs}

\begin{thebibliography}{10}\itemsep=-1pt

\bibitem{alvarez2016learning}
Jose Alvarez and Mathieu Salzmann.
\newblock Learning the number of neurons in deep networks.
\newblock In {\em Advances in Neural Information Processing Systems (NeurIPS)},
  2016.

\bibitem{arpit2017closer}
Devansh Arpit et~al.
\newblock A closer look at memorization in deep networks.
\newblock In {\em International Conference on Machine Learning (ICML)}, 2017.

\bibitem{baldi1989neural}
Pierre Baldi and Kurt Hornik.
\newblock Neural networks and principal component analysis: Learning from
  examples without local minima.
\newblock {\em Neural networks}, 2(1), 1989.

\bibitem{benedetto2003finite}
John Benedetto and Matthew Fickus.
\newblock Finite normalized tight frames.
\newblock {\em Advances in Computational Mathematics}, 18(2-4), 2003.

\bibitem{bruckstein2008uniqueness}
Alfred~M. Bruckstein et~al.
\newblock On the uniqueness of nonnegative sparse solutions to underdetermined
  systems of equations.
\newblock {\em IEEE Transactions on Information Theory}, 54(11), 2008.

\bibitem{cortes1995support}
Corinna Cortes and Vladimir Vapnik.
\newblock Support-vector networks.
\newblock {\em Machine learning}, 20(3), 1995.

\bibitem{denil2013predicting}
Misha Denil et~al.
\newblock Predicting parameters in deep learning.
\newblock In {\em Advances in Neural Information Processing Systems (NeurIPS)},
  2013.

\bibitem{donoho2006compressed}
David~L. Donoho.
\newblock Compressed sensing.
\newblock {\em IEEE Transactions on Information Theory}, 52(4), 2006.

\bibitem{donoho2003optimally}
David~L. Donoho and Michael Elad.
\newblock Optimally sparse representation in general (nonorthogonal)
  dictionaries via l1 minimization.
\newblock {\em Proceedings of the National Academy of Sciences}, 100(5), 2003.

\bibitem{donoho2005stable}
David~L. Donoho, Michael Elad, and Vladimir~N. Temlyakov.
\newblock Stable recovery of sparse overcomplete representations in the
  presence of noise.
\newblock {\em IEEE Transactions on Information Theory}, 52(1), 2005.

\bibitem{elad2010sparse}
Michael Elad.
\newblock {\em Sparse and redundant representations: from theory to
  applications in signal and image processing}.
\newblock Springer Science \& Business Media, 2010.

\bibitem{elsken2019neural}
Thomas Elsken, Jan~Hendrik Metzen, and Frank Hutter.
\newblock Neural architecture search: A survey.
\newblock {\em Journal of Machine Learning Research (JMLR)}, 20(55), 2019.

\bibitem{he2016deep}
Kaiming He, Xiangyu Zhang, Shaoqing Ren, and Jian Sun.
\newblock Deep residual learning for image recognition.
\newblock In {\em Conference on Computer Vision and Pattern Recognition
  (CVPR)}, 2016.

\bibitem{he2017channel}
Yihui He et~al.
\newblock Channel pruning for accelerating very deep neural networks.
\newblock In {\em Conference on Computer Vision and Pattern Recognition
  (CVPR)}, 2017.

\bibitem{huang2017densely}
Gao Huang, Zhuang Liu, Kilian Weinberger, and Laurens van~der Maaten.
\newblock Densely connected convolutional networks.
\newblock In {\em Conference on Computer Vision and Pattern Recognition
  (CVPR)}, 2017.

\bibitem{ioffe2015batch}
Sergey Ioffe and Christian Szegedy.
\newblock Batch normalization: Accelerating deep network training by reducing
  internal covariate shift.
\newblock In {\em International Conference on Machine Learning (ICML)}, 2015.

\bibitem{kovavcevic2008introduction}
Jelena Kova{\v{c}}evi{\'c}, Amina Chebira, et~al.
\newblock An introduction to frames.
\newblock {\em Foundations and Trends in Signal Processing}, 2(1), 2008.

\bibitem{krizhevsky2012imagenet}
Alex Krizhevsky et~al.
\newblock Imagenet classification with deep convolutional neural networks.
\newblock In {\em Advances in Neural Information Processing Systems (NeurIPS)},
  2012.

\bibitem{krizhevsky2009learning}
Alex Krizhevsky and Geoffrey Hinton.
\newblock Learning multiple layers of features from tiny images.
\newblock Technical report, University of Toronto, 2009.

\bibitem{lecun1998gradient}
Yann LeCun, L{\'e}on Bottou, Yoshua Bengio, and Patrick Haffner.
\newblock Gradient-based learning applied to document recognition.
\newblock {\em Proceedings of the IEEE}, 86(11), 1998.

\bibitem{li2018densepack}
Yixuan Li.
\newblock Tensorflow densenet.
\newblock \url{https://github.com/YixuanLi/densenet-tensorflow}, 2018.

\bibitem{moustapha2017parseval}
Cisse Moustapha, Bojanowski Piotr, Grave Edouard, Dauphin Yann, and Usunier
  Nicolas.
\newblock Parseval networks: Improving robustness to adversarial examples.
\newblock In {\em International Conference on Machine Learning (ICML)}, 2017.

\bibitem{murdock2018deep}
Calvin Murdock, MingFang Chang, and Simon Lucey.
\newblock Deep component analysis via alternating direction neural networks.
\newblock In {\em European Conference on Computer Vision (ECCV)}, 2018.

\bibitem{neyshabur2017exploring}
Behnam Neyshabur, Srinadh Bhojanapalli, David McAllester, and Nati Srebro.
\newblock Exploring generalization in deep learning.
\newblock In {\em Advances in Neural Information Processing Systems (NeurIPS)},
  2017.

\bibitem{novak2018sensitivity}
Roman Novak et~al.
\newblock Sensitivity and generalization in neural networks: an empirical
  study.
\newblock In {\em International Conference on Learning Representations (ICLR)},
  2018.

\bibitem{papyan2017convolutional}
Vardan Papyan, Yaniv Romano, and Michael Elad.
\newblock Convolutional neural networks analyzed via convolutional sparse
  coding.
\newblock {\em Journal of Machine Learning Research (JMLR)}, 18(83), 2017.

\bibitem{romano2018adversarial}
Yaniv Romano, Aviad Aberdam, Jeremias Sulam, and Michael Elad.
\newblock Adversarial noise attacks of deep learning architectures-stability
  analysis via sparse modeled signals.
\newblock {\em Journal of Mathematical Imaging and Vision}, 2018.

\bibitem{simonyan2014very}
Karen Simonyan and Andrew Zisserman.
\newblock Very deep convolutional networks for large-scale image recognition.
\newblock In {\em International Conference on Learning Representations (ICLR)},
  2015.

\bibitem{sulam2019multi}
Jeremias Sulam, Aviad Aberdam, Amir Beck, and Michael Elad.
\newblock On multi-layer basis pursuit, efficient algorithms and convolutional
  neural networks.
\newblock {\em Pattern Analysis and Machine Intelligence (PAMI)}, 2019.

\bibitem{tan2019efficientnet}
Mingxing Tan and Quoc Le.
\newblock Efficientnet: Rethinking model scaling for convolutional neural
  networks.
\newblock In {\em International Conference on Machine Learning (ICML)}, 2019.

\bibitem{vapnik1971uniform}
Vladimir~N. Vapnik and A.~Ya. Chervonenkis.
\newblock On the uniform convergence of relative frequencies of events to their
  probabilities.
\newblock {\em Theory of Probability and Its Applications}, XVI(2), 1971.

\bibitem{welch1974lower}
Lloyd Welch.
\newblock Lower bounds on the maximum cross correlation of signals.
\newblock {\em IEEE Transactions on Information theory}, 20(3), 1974.

\bibitem{wu2016tensorpack}
Yuxin Wu et~al.
\newblock Tensorpack.
\newblock \url{https://github.com/tensorpack/}, 2016.

\bibitem{zhang2017understanding}
Chiyuan Zhang, Samy Bengio, Moritz Hardt, Benjamin Recht, and Oriol Vinyals.
\newblock Understanding deep learning requires rethinking generalization.
\newblock In {\em International Conference on Learning Representations (ICLR)},
  2017.

\end{thebibliography}
}

\end{document}